\documentclass[journal]{IEEEtran}

\usepackage{times}
\usepackage{epsfig}
\usepackage{graphicx}
\usepackage{amsmath}
\usepackage{amssymb}
\usepackage{multirow}
\usepackage{subfigure}
\usepackage{color}
\usepackage{algorithm}
\usepackage{booktabs}
\usepackage{algorithmic}
\usepackage[pagebackref=true,breaklinks=true,letterpaper=true,colorlinks,bookmarks=false]{hyperref}

\newcommand{\tabincell}[2]{\begin{tabular}{@{}#1@{}}#2\end{tabular}}

\begin{document}
\title{Distilling a Powerful Student Model via \\Online Knowledge Distillation}

\author{Shaojie Li,
        Mingbao Lin,
        Yan Wang, 
        Yongjian Wu, 
        Yonghong Tian,~\IEEEmembership{Fellow,~IEEE}, \\
        Ling Shao,~\IEEEmembership{Fellow,~IEEE},
        Rongrong Ji,~\IEEEmembership{Senior Member,~IEEE}% <-this % stops a space
\IEEEcompsocitemizethanks{
\IEEEcompsocthanksitem S. Li is with the Media Analytics and Computing Laboratory, Department of Artificial Intelligence, School of Informatics, Xiamen University, Xiamen 361005, China.\protect
\IEEEcompsocthanksitem M. Lin is with the  Media Analytics and Computing Laboratory, Department of Artificial Intelligence, School of Informatics, Xiamen University, Xiamen 361005, China, also with the Youtu Laboratory, Tencent, Shanghai 200233, China. \protect
\IEEEcompsocthanksitem Y. Wang is with Pinterest, USA.\protect
\IEEEcompsocthanksitem Y. Wu is with Youtu Laboratory, Tencent, Shanghai 200233, China.\protect
\IEEEcompsocthanksitem Y. Tian is with the School of Electronics Engineering and Computer Science, Peking University, Beijing 100871, China.\protect 
\IEEEcompsocthanksitem L. Shao is with the Inception Institute of Artificial Intelligence, Abu Dhabi, United Arab Emirates, and also with the Mohamed bin Zayed University of Artificial Intelligence, Abu Dhabi, United Arab Emirates. \protect
\IEEEcompsocthanksitem R. Ji (Corresponding Author) is with the Media Analytics and Computing
Laboratory, Department of Artificial Intelligence, School of Informatics, Xia-
men University, Xiamen 361005, China, also with Institute of Artificial Intelli-
gence, Xiamen University, Xiamen 361005, China (e-mail: rrji@xmu.edu.cn).
}% <-this % stops a space
\thanks{Manuscript received April 19, 2005; revised August 26, 2015.}}


\markboth{IEEE TRANSACTIONS ON NEURAL NETWORKS AND LEARNING SYSTEMS}%
{Shell \MakeLowercase{\textit{et al.}}: Bare Demo of IEEEtran.cls for IEEE Journals}



\maketitle

%%%%%%%%% ABSTRACT
\begin{abstract}
Existing online knowledge distillation approaches either adopt the student with the best performance or construct an ensemble model for better holistic performance. However, the former strategy ignores other students' information, while the latter increases the computational complexity during deployment. In this paper, we propose a novel method for online knowledge distillation, termed FFSD, which comprises two key components: Feature Fusion and Self-Distillation, towards solving the above problems in a unified framework. Different from previous works, where all students are treated equally, the proposed FFSD splits them into a leader student and a common student set. Then, the feature fusion module converts the concatenation of feature maps from all common students into a fused feature map. The fused representation is used to assist the learning of the leader student. To enable the leader student to absorb more diverse information, we design an enhancement strategy to increase the diversity among students. Besides, a self-distillation module is adopted to convert the feature map of deeper layers into a shallower one. Then, the shallower layers are encouraged to mimic the transformed feature maps of the deeper layers, which helps the students to generalize better. After training, we simply adopt the leader student, which achieves superior performance, over the common students, without increasing the storage or inference cost. Extensive experiments on CIFAR-100 and ImageNet demonstrate the superiority of our FFSD over existing works. The code is available at \url{https://github.com/SJLeo/FFSD}.
\end{abstract}

%%%%%%%%% BODY TEXT
\section{Introduction}\label{introduction}

\IEEEPARstart{D}{eep}  neural networks (DNNs) have achieved unprecedented success in various visual tasks.
%, including image classification~~\cite{he2016deep,szegedy2015going}, object detection~~\cite{girshick2014rich,ren2015faster}, and image generation~~\cite{radford2015unsupervised,zhu2017unpaired}. 
Nevertheless, their extensive memory and computational requirements hinder their deployment in resource-limited devices. Several methods have been developed to derive a light-weight model with negligible performance compromise.
Examples include network pruning~\cite{han2015learning,he2019filter}, parameter quantization~\cite{rastegari2016xnor,jacob2018quantization}, low-rank decomposition~\cite{denton2014exploiting,zhang2015efficient} and knowledge distillation~\cite{hinton2015distilling,romero2014fitnets}.

\begin{figure}
	\centering
% 	\subfigure[Traditional Knowledge Distillation]{
% 		\begin{minipage}{\linewidth}
% 	        \subfigure[First Stage]{\includegraphics[width=0.46\linewidth]{plot/KD_stage1.pdf}}
%             \subfigure[Second Stage]{\includegraphics[width=0.46\linewidth]{plot/KD_stage2.pdf}}
% 		\end{minipage}
% 	}
    \subfigure[Traditional Knowledge Distillation]{
    % 		\begin{minipage}{0.24\textwidth}%[b]%{0.2\textwidth}
    			\includegraphics[width=0.85\linewidth]{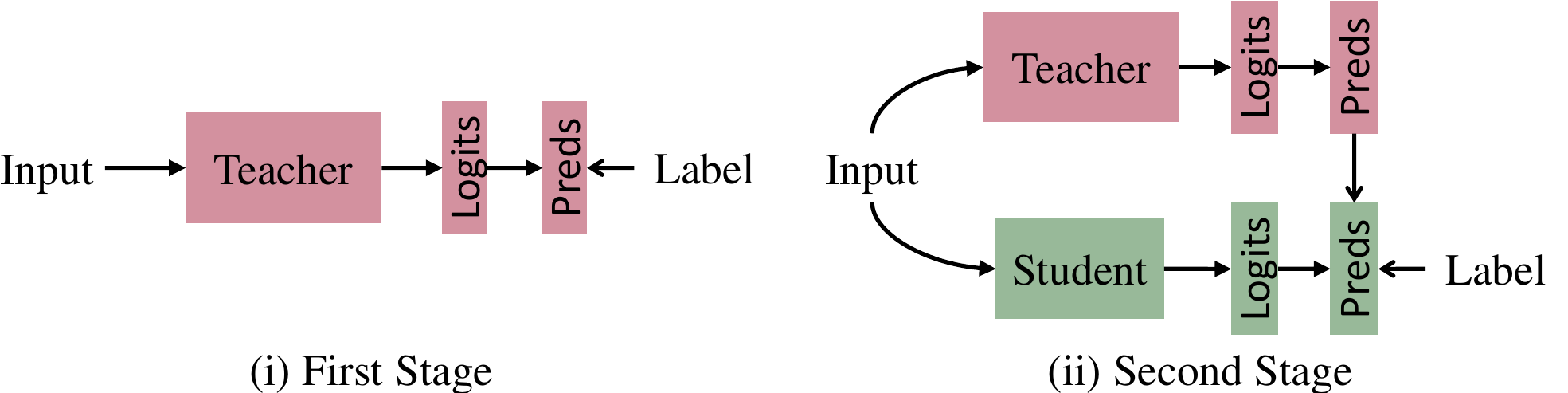} 
    % 		\end{minipage}
	}
	\subfigure[Mutual Learning]{
% 		\begin{minipage}{0.24\textwidth}%[b]%{0.2\textwidth}
			\includegraphics[width=0.42\linewidth]{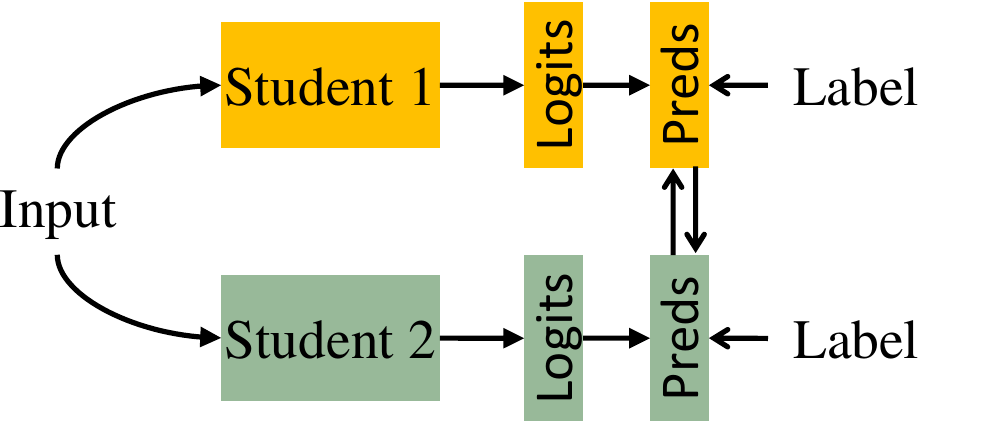} 
% 		\end{minipage}
	}%
	\subfigure[Ensemble Learning]{
% 		\begin{minipage}{0.24\textwidth}%[b]%{0.2\textwidth}
			\includegraphics[width=0.42\linewidth]{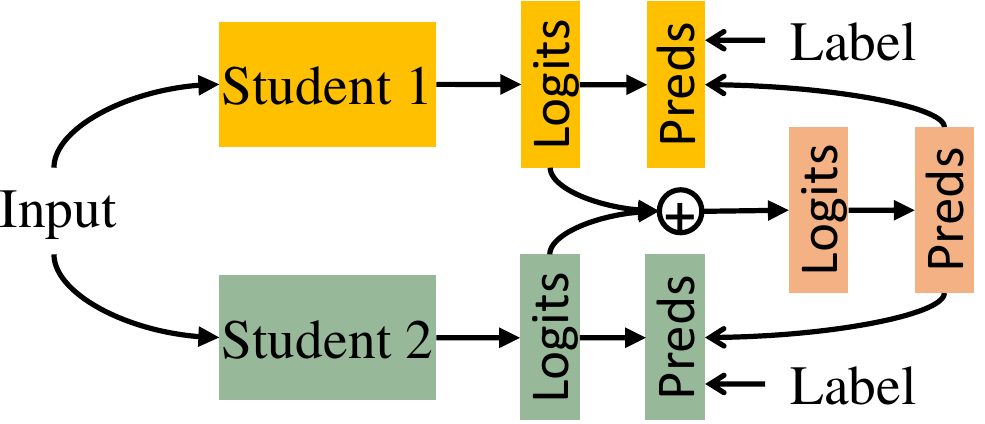} 
% 		\end{minipage}
	}
	\caption{Three types of knowledge distillation. (a) displays the two-stage optimization of distillation, which has to pre-train a large-scale teacher model. (b) and (c) outline the online distillation using either mutual learning or ensemble learning, which does not involve a teacher model.} 
	\vspace{-1.5em}
	\label{comparison}
\end{figure}

Among them, knowledge distillation has received particular attention, transfering knowledge from a high-capacity teacher~\cite{hinton2015distilling, romero2014fitnets, komodakis2017paying}, or an online ensemble~\cite{zhang2018deep, song2018collaborative, zhu2018knowledge}, to a student model.
As illustrated in Fig.\,\ref{comparison}(a), 
Traditional knowledge distillation methods use a two-stage optimization where a cumbersome teacher network has to be trained in advance in order to yield a high-capacity model, which then serves as supervision information to guide the training of a light-weight student network.
Though progress has been made, these methods heavily rely on an appropriate teacher model. As stressed in~\cite{jin2019knowledge, cho2019efficacy}, it is difficult to choose a suitable teacher model for the student model.

% these models require significant training time due to their heavy dependence on a large-scale teacher model.

This has motivated the community to simplify the training procedure by exploring online knowledge distillation~\cite{zhang2018deep,zhu2018knowledge}, where a collection of student models are trained simultaneously in a collaborative manner without the involvement of a teacher model.
As shown in Fig.\,\ref{comparison}, 
existing online knowledge distillation can be implemented by either mutual learning~\cite{zhang2018deep, chung2020feature, zhang2020amln} or ensemble learning~\cite{zhu2018knowledge, song2018collaborative, chen2020online, guo2020online, kim2019feature}.
The former aligns the soft outputs of all students so as to allow message passing among them.
Then, the student model with the optimal performance is adopted as the final model.
However, the message passing does not guarantee that one single student will carry all the information of the ensemble.
This limits the distillation performance.
In contrast, the latter constructs a virtual teacher by ensembling the outputs of all the students, which are then distilled back to foster each student.
Nevertheless, this group has to retain all the student models so as to ensemble their output logits together to pursue better performance, which significantly increases the memory consumption and computational complexity since all models have to be stored in disks and evaluated during inference.
Thus, online knowledge distillation remains an open problem.
%On one hand, we pursue information from all students to derive a more powerful student model. On the other hand, we aim to free from resource overhead in the deployment.

\begin{figure}
\begin{center}
\includegraphics[width=0.9\linewidth]{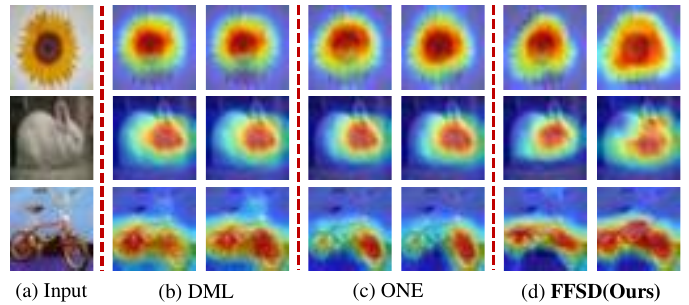}
\end{center}
\vspace{-1.0em}
\caption{Feature map visualization of the two student models using ResNet-32 on CIFAR-100. Feature maps of student models are highly similar in DML~\cite{zhang2018deep} and ONE~\cite{zhu2018knowledge}. Our FFSD can learn diversified feature maps.}
\vspace{-1.5em}
\label{fig:diversity_motivation}
\end{figure}

In this paper, we present a novel online knowledge distillation method, introducing two key modules, \emph{i.e.}, feature fusion and self-distillation (FFSD), in order to solve the above problems in a unified framework. Specifically, we construct a common student set and a leader student, which share the same network architecture.
%The common student set contains at least two student models, all student models and the leader student model use the same network architecture. 
The common student set is trained through mutual learning.
In order to take full advantage of the rich information of all common students in the set, we design a feature fusion module similar to an autoencoder to fuse the output feature maps from all common students, and then distill it to the leader student.
First, it encodes the common students' output feature maps into a meaningful compact feature map with the same size as the leader student, and then the leader student is encouraged to mimic this fused feature map.
Meanwhile, the output feature map of the leader student is decoded back to match the concatenated feature maps of all common students to ensure the effectiveness of the feature map learned by the leader student.
However, as shown in Fig.\,\ref{fig:diversity_motivation}, the activation feature maps of mutual learning and ensemble learning are highly overlapped, which means that the common students all pay more attention to exactly the same area of the input image. 
The leader student thus cannot obtain additional information from the fused feature map.
To address this issue, we devise a strategy to enhance the feature diversity of the common student models in the set.
Specifically, we construct a diverse feature map by shifting the focus of the output feature map of a student.
The diverse feature map is further distilled to other students.
In addition, we design a self-distillation module that converts high-order supervision information (the fused feature map and the diverse feature map) to low-order supervision information for shallower network layers.
Self-distillation facilitates the flow of supervision information from deeper to shallower network layers.
It thus enriches the supervision information in the training process.
After training, only the leader student is retained for deployment, as it achieves superior performance over all other students.
%We empirically demonstrate the effectiveness of our FFSD with extensive experiments on the CIFAR-100~\cite{krizhevsky2009learning} and ImageNet~\cite{russakovsky2015imagenet} datasets. 

Our contributions are summarized as follows:
\begin{itemize}
    \item A novel online knowledge distillation method, FFSD, is proposed. We design a feature fusion module equipped with a diversity enhancement strategy to integrate the knowledge of students and distill it to the leader student, which improves the generality of the final model.
    \item A self-distillation module is proposed to convert high-order supervision information to low-order cues for shallower network layers, which provides richer information for model training.
    \item Extensive experiments on two benchmarks demonstrate the effectiveness of our FFSD.
\end{itemize}

\section{Related Work}\label{related}

\textbf{Traditional Knowledge Distillation.} Traditional distillation works transfer knowledge from a cumbersome teacher model to a light-weight student model. As such, a large-scale model has to be trained in advance, based on which various knowledge definitions and transfer strategies are proposed to boost the performance of the student model. The pioneering work~\cite{hinton2015distilling} performs knowledge representation of a teacher model using the softmax output layer, which converts the logit into a soft probability with a temperature parameter. Following this, a large number of works proposed new forms of knowledge, such as output logits~\cite{yuan2020revisiting, ding2021knowledge}, intermediate feature maps~\cite{romero2014fitnets, chen2021distilling, chen2021cross}, attention maps~\cite{komodakis2017paying}, second-order statistics~\cite{yim2017gift}, contrastive features~\cite{tian2019contrastive,xu2020knowledge}, or structured knowledge~\cite{park2019relational,liu2019knowledge, passalis2020heterogeneous}. Another group of methods focus on transfer strategies so as to enable the student model to inherit knowledge from the teacher model. An intuitive solution is to use the Kullback-Leibler divergence or $\ell_p$-loss when the knowledge falls on the soft logit~\cite{hinton2015distilling,li2017learning} or intermediate representation~\cite{romero2014fitnets,komodakis2017paying}. Beyond that, Wang \emph{et al}.~\cite{wang2018kdgan} utilized the adversarial training scheme in generative adversarial networks (GANs)~\cite{goodfellow2014generative} to transfer knowledge. Jang \emph{et al}.~\cite{jang2019learning} considered meta-learning to selectively transfer knowledge. In~\cite{liu2020search}, a reinforcement learning based architecture-aware distillation was proposed to pass the structural knowledge to the student. Recently, there are some works~\cite{li2020gan, li2020learning, jin2021teachers, liu2021content} that used knowledge distillation for GAN compression to ensure effective compression. The surveys~\cite{gou2021knowledge, wang2021knowledge} summarized the development of knowledge distillation in recent years.

\textbf{Online Knowledge Distillation.} Online knowledge distillation has emerged as an alternative that eliminates the dependency on the teacher model. It builds knowledge distillation based on a collection of student models that collaborate through simultaneous training. To this end, Zhang \emph{et al}.~\cite{zhang2018deep} proposed a deep mutual learning strategy where pair-wise students are encouraged to learn from each other by a mimicry loss based on the Kullback-Leibler divergence. Chen \emph{et al}.~\cite{chen2020online} performed two-level distillation by training multiple auxiliary peers and one group leader, separately. The former aims to boost peer diversity, while the latter transfers knowledge from an ensemble of auxiliary peers to the group leader. In~\cite{chung2020feature}, online knowledge distillation was built at a feature-map level using the adversarial training framework. Kim \emph{et al}.~\cite{kim2019feature} fused the intermediate representations of subnetworks, passing the result to an auxiliary classifier. Then, the knowledge from the auxiliary classifier is delivered back to each subnetwork for mutual teaching. In~\cite{zhu2018knowledge,song2018collaborative,guo2020online}, all student branches are ensembled to construct a stronger teacher model, which is in turn distilled back to the students to enhance the model learning.
EnD2~\cite{malinin2020ensemble} distilled the distribution of the predictions of the ensemble into a single model. It enables a single model to retain both the performance of the ensemble as well as the ability of uncertainty estimation.
EnD2 also performs diversity enhancement to capture uncertain information. However, we enhance the diversity among common students so that student leaders can obtain higher benefits from feature fusion.

%AMLN~\cite{zhang2020amln} introduced process-driven learning beyond outcome-driven learning for augmented online knowledge distillation. DCM~\cite{yao2020knowledge} connected knowledged representation learning with deep supervision methodology and introduces dense cross-layer bidirectional knowledge distillation method. 

\textbf{Self-Distillation.} Self-distillation, originally proposed by Furlanello \emph{et al}.~\cite{furlanello2018born}, has received a great deal of attention recently, due to its distillation of knowledge within the network itself without the aid of other models. Augmentation based works~\cite{lee2019rethinking, xu2019data} focus on self-distillation via data augmentation of the input images. Hou \emph{et al}.~\cite{hou2019learning} and Zhang \emph{et al}.~\cite{zhang2019your} distilled deeper parts of the network as the conceptual teacher model to guide the learning of shallower modules. Li \emph{et al}.~\cite{yuan2020revisiting} revisited knowledge distillation as a type of learned label smoothing regularization, and accordingly proposed a novel teacher-free knowledge distillation framework where the student model learns from itself or a manually designed regularization distribution. In addition, some works have achieved superior performance by applying self-distillation on object detection~\cite{huang2020comprehensive, zheng2021se} and super-resolution~\cite{wang2021towards}.

\section{The Proposed Method}
%In this section, we elaborate the overall process of our proposed method. In Section 3.1, we

\subsection{Preliminaries}
Consider a group of $n + 1$ student models $\mathbf{G} = \{ \mathbf{S}_i \}_{i=0}^n$, all of which share the same network structure and consist of $L$ convolutional layers. For each model $\mathbf{S}_i$, we denote the output of the feature maps in the $l$-th layer as $\mathbf{F}_i^l \in \mathbb{R}^{C^l\times H^l\times W^l}$, where $C^l,H^l,W^l$ denote the channel number, height and width of a feature map, respectively. Besides, given a labeled dataset $\mathcal{D}=\left\{\left(\mathbf{x}, \mathbf{y}\right)\right\}$ with $K$ classes, the logit produced by student $\mathbf{S}_i$ is denoted as $\mathbf{z}_i \in \mathbb{R}^{K}$.

Then, the prediction probability of the softmax layer is represented by $\mathbf{p}_i$, with the $k$-th class computed as
\begin{equation}
\label{equation:temperature_softmax}
\mathbf{p}_i^k=\dfrac{\exp(\mathbf{z}_i^k/T)}{\sum_{k=1}^K\exp(\mathbf{z}_i^k/T)},
\end{equation}
where $T \ge 1$ is the temperature parameter used to soften the output probability. When $T=1$, it degenerates to the original softmax output. For ease of representation, we consider $\mathbf{p}_i$ as having temperature $T = 1$; otherwise, we rewrite it as $\hat{\mathbf{p}}_i$.

Existing literature encourages all students $\mathbf{S}_i \in \mathbf{G}$ to learn from each other. Then, the resulting model for deployment falls on the optimal student or the ensemble one. As discussed in Sec.\,\ref{introduction}, the former ignores the efficacy of other students, while the latter increases the resource burden. Differently, in this paper, we innovatively propose to regard $\mathbf{S}_0$ as the leader student and the remaining $\hat{\mathbf{G}} = \mathbf{G} - \{\mathbf{S}_0\}$ as a common student set. Then, students in $\hat{\mathbf{G}}$ learn in a collaborative manner, while the leader student $\mathbf{S}_0$ is responsible for learning the knowledge from the common students. All students mentioned below represent the common students in the set $\hat{\mathbf{G}}$.% and not the leader student.

Following \cite{zhang2018deep}, we first define the training objective $\mathcal{L}_{\text{base}}$ for the collaborative learning of the students as
\begin{equation}
\label{equation:baseloss}
\mathcal{L}_{\text{base}}=\mathcal{L}_{CE}(\mathbf{y}, \mathbf{p}_i)+ \\
                         T^2\sum_{j=1,j\neq i}^{n}\mathcal{L}_{KL}(\hat{\mathbf{p}_j}||\hat{\mathbf{p}_i}),
\end{equation}
where $\mathcal{L}_{CE}$ is the cross-entropy loss between the one-hot ground-truth label $\mathbf{y}$ and the prediction $\mathbf{p}_i$. The $\mathcal{L}_{KL}$ from $\hat{\mathbf{p}_i}$ to $\hat{\mathbf{p}_j}$ is computed as
\begin{equation}
\mathcal{L}_{KL}(\hat{\mathbf{p}}_j||\hat{\mathbf{p}}_i)=\sum_{k=1}^{K} \hat{\mathbf{p}}_{j}^{k} \log \dfrac{\hat{\mathbf{p}}_{j}^{k}}{\hat{\mathbf{p}}_{i}^{k}}.
\end{equation}

We multiply the $\mathcal{L}_{KL}$ with $T^2$ because the gradients produced by the soft predictions are scaled by $1/T^2$.
\begin{figure}
\begin{center}
\includegraphics[width=0.8\linewidth, height=0.3\columnwidth]{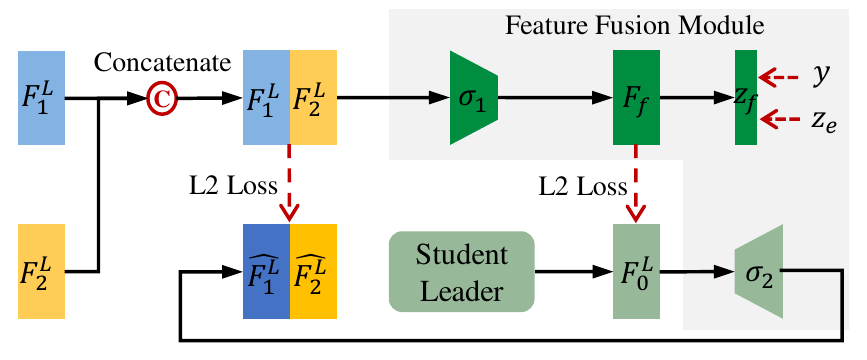}
\end{center}
\vspace{-1.0em}
\caption{Feature fusion for feature learning of the leader student. The learning scheme uses an autoencoder framework that forces the last-layer output feature map $\mathbf{F}_0^L$ of the leader student to mimic the fused feature map $\mathbf{F}_f$.}
\vspace{-1.5em}
\label{fig:feature_fusion_module}
\end{figure}

\subsection{Feature Fusion}\label{FeatureFusion}
To maximize the usage of the students' information, we design a feature fusion module in Fig.\,\ref{fig:feature_fusion_module}, to fuse the feature maps of all students. The result is then distilled to strengthen the capacity of the leader student $\mathbf{S}_0$.

Specifically, we first concatenate the feature maps in the $L$-th convolutional layers of all common students, \emph{i.e.}, $\{ \mathbf{F}_i^L \}_{i=1}^n$, the result of which is denoted as $\mathbf{F}_e$. The fusion module encodes the concatenated feature maps into a meaningful compact feature map $\mathbf{F}_f$ with the same size as the leader student. This fused feature map is then passed into a fusion classifier supervised by the ground-truth labels and the ensemble logit of students $\mathbf{z}_e = \frac{1}{n}\sum_{i=1}^n \mathbf{z}_i$~\cite{zhang2018deep, song2018collaborative}. We further denote the output logit of the fusion classifier as $\mathbf{z}_f$. By transferring these logits into prediction probabilities using Eq.\,(\ref{equation:temperature_softmax}), the training objective for the fusion classifier is computed as
\begin{equation}
\label{equation:fusion_module_loss}
    \mathcal{L}_{\text{fusion}} = \mathcal{L}_{CE}(\mathbf{y}, \mathbf{p}_f)+ \\
                          T^2\mathcal{L}_{KL}(\hat{\mathbf{p}}_e||\hat{\mathbf{p}}_f).
\end{equation}

We aim to transfer the high-quality information from the fused feature map to the leader student. To this end, we encourage the last-layer output of the leader student $\mathbf{F}_0^L$ to learn from the fused feature map. Meanwhile, the output feature map of the leader student is decoded back to match the concatenated feature maps of all the students to ensure the effectiveness of the feature map learned by the leader student.
%It ensures the effectiveness of the learned feature map that the output feature map of the leader student is decoded back to match the concatenated feature maps. 
Hence, we can derive our optimization objective for the output feature map of the leader student as follows
\begin{equation}
\label{equation:feature_loss}
    \mathcal{L}_{\mathbf{F}_0^L} =
    \bigg\|\dfrac{\mathbf{F}_0^L}{\|\mathbf{F}_0^L\|_2}-\dfrac{\mathbf{F}_f}{\|\mathbf{F}_f\|_2}\bigg\|_2 +
    \bigg\|\dfrac{\sigma(\mathbf{F}_0^L)}{\|\sigma(\mathbf{F}_0^L)\|_2}-\dfrac{\mathbf{F}_e}{\|\mathbf{F}_e\|_2}\bigg\|_2,
\end{equation}
where $\|\cdot\|_2$ is the $\ell_2$-norm and $\sigma(\cdot)$ aligns the channel dimension of $\mathbf{F}_0^L$ to $\mathbf{F}_e$.

\begin{figure*}
	\centering
	\subfigure[Training Framework]{
			\includegraphics[width=0.4\linewidth]{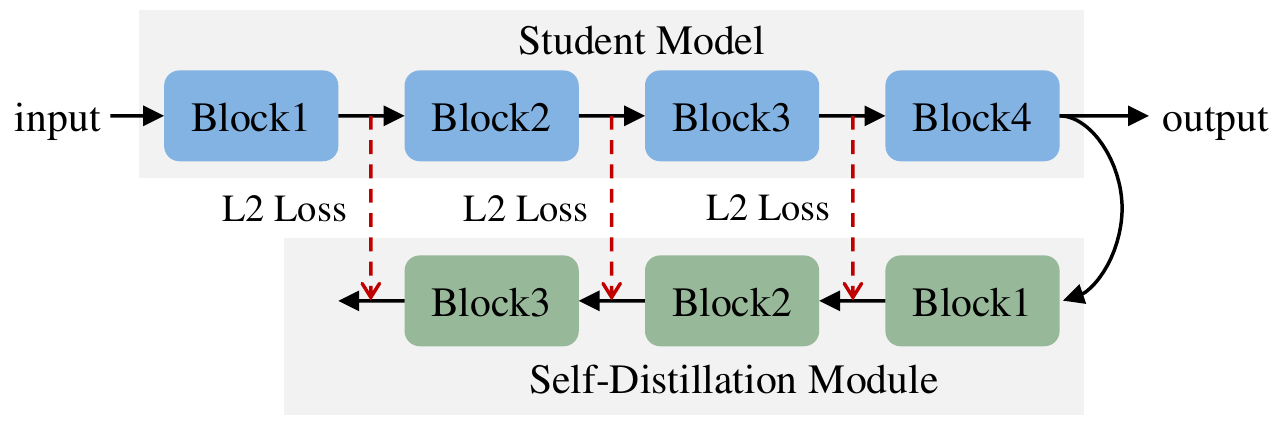} 
			\label{fig:self_distillation_training}
			}
	\subfigure[Inference Framework]{
			\includegraphics[width=0.46\linewidth]{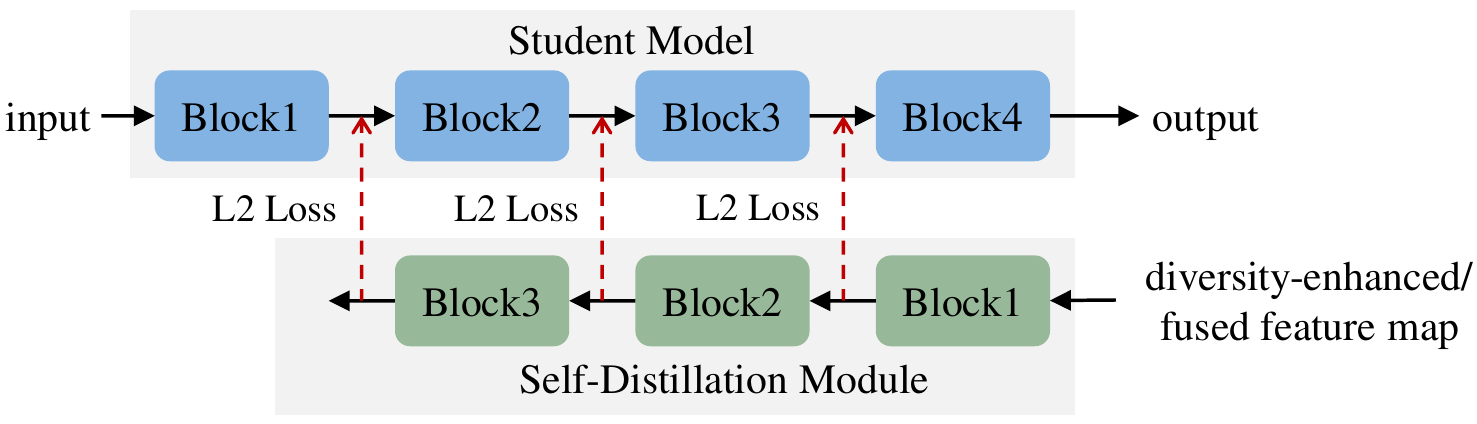} 
			\label{fig:self_distillation_inference}
	}
	\vspace{-0.5em}
	\caption{The working flows of our proposed self-distillation module in the training stage (a) and inference stage (b).}
	\vspace{-1.5em}
	\label{fig:self_distillation}
\end{figure*}

The layer-wise feature amalgamation is performed from multiple teachers in~\cite{shen2019amalgamating}, but we only perform feature fusion in the last layer, which is then distilled to the shallower layers. During fusion process, we set up an additional classifier to supervise the quality of the fused features, which is not available in~\cite{shen2019amalgamating}. Also, as shown in Fig.\,\ref{fig:diversity_motivation}, the output representations in mutual learning~\cite{zhang2018deep} or ensemble learning~\cite{zhu2018knowledge} tend to be unified, preventing the leader student from additional information of the feature fusion. Thus, it is necessary to diversify the student outputs, which is ignored in~\cite{shen2019amalgamating}. A intuitive solution can resort to minimizing the negative reconstruction error on the intermediate outputs~\cite{romero2014fitnets,luo2017thinet,he2017channel} as
\begin{equation}
\label{equation:loss_dir}
\mathcal{L}_{\text{div}}= - \dfrac{1}{L} \sum_{i = 1, j = 1, i\neq j}^n\sum_{l=1}^L\|\mathbf{F}_i^l-\mathbf{F}_j^l\|_2.
\end{equation}

However, Eq.\,(\ref{equation:loss_dir}) raises some issues: (1) \textbf{Significant computational complexity.} A total of $n(n-1) L$ $\ell_2$-norm distances have to be computed. (2) \textbf{Task independence.} The $\ell_2$-norm loss is to reduce the overall error, which will shift the attention of feature maps to a task-independent position. (3) \textbf{Attention inconsistency.} The per-layer loss is calculated separately, ignoring the coherence of attentions across different layers.

To reduce the diversity computation, we propose to train the first student $\mathbf{S}_1$ using Eq.\,(\ref{equation:baseloss}), where student $\mathbf{S}_i (i > 1)$ only performs diversity enhancement calculation with student $\mathbf{S}_{i - 1}$. Thus, our diversity enhancement learning transfers each student's knowledge to the next peer student in a one-way chain manner. However, diversifying feature maps from all layers still incurs high complexity. Fortunately, we observe using only the last layer of each residual block for ResNet~\cite{he2016deep} can perform well. We denote the selected feature maps of student $\mathbf{S}_i$ as $\{ \mathbf{F}_i^m \}_{m=1}^M \in \{\mathbf{F}_i^l\}_{l=1}^L$ for diversity. Note $F_i^M=F_i^L$.
The diversity is essentially enhanced to enable the attention of students to concentrate on different image positions. We first extract the attention for each feature map $\mathbf{F}_i^m$~\cite{komodakis2017paying} as
\begin{equation}
\label{equation:attention_calculation}
\mathbf{A}_{i}^{m}=\sum_{c=1}^{C^m}|\mathbf{F}_i^m(c,:,:)|^2,
\end{equation}
where $C^m$ denotes the channel number of the $m$-th layer. Then, the diversity enhancement attention map $\bar{\mathbf{A}}_i^m$ is
\begin{equation}
\label{equation:diveristy_enhancment}
\bar{\mathbf{A}}_i^m=(\dfrac{P}{2}-\mathbf{A}_i^m)\times sign(\mathbf{A}_i^m-t)+\dfrac{P}{2},
\end{equation}
where $P=\|\mathbf{A}_i^m\|_2$ and $t$ is set as the $\frac{H^m\times W^m}{3}$-th smallest number in $\mathbf{A}_i^m$. The goal of Eq.\,(\ref{equation:diveristy_enhancment}) is to shift attention $\mathbf{A}_i^m$ to the slightly weaker areas, while maintaining the attention value of task-independent areas. 
Specially, $sign(\mathbf{A}_i^m-t)$ is used to determine whether the area is task-dependent. When the area is task-independent ($\mathbf{A}_i^m<t, sign(\mathbf{A}_i^m-t)=-1$), the value of $\bar{\mathbf{A}}_i^m$ is equal to $\mathbf{A}_i^m$, and $P-A_i^m$, otherwise.
We shift attention to the slightly weaker areas by changing the activation value of the task-dependent area into $P-A_i^m$. Our enhancement for $\mathbf{S}_i(i>1)$ to replace Eq.\,(\ref{equation:loss_dir}) is
\begin{equation}
\label{equation:diversity_loss}
\mathcal{L}_{\text{div}}=\sum_{m=1}^{M}\bigg\|\dfrac{\mathbf{A}_i^m}{\|\mathbf{A}_i^m\|_2}-\dfrac{\bar{\mathbf{A}}_{i-1}^m}{\|\bar{\mathbf{A}}_{i-1}^m\|_2}\bigg\|_2.
\end{equation}

In what follows, we detail our self-distillation to solve the problem of attention inconsistency.

% \begin{figure}
% \begin{center}
% \includegraphics[width=0.8\linewidth]{plot/self_distllation_module.pdf}
% \end{center}
% \vspace{-1.0em}
% \caption{Training scheme of the self-distillation module.}
% \vspace{-0.5em}
% \label{fig:self_distillation}
% \end{figure}

\subsection{Self-Distillation}\label{selfdistillation}
This section introduces a novel self-distillation module to convert high-order information to low-order information for shallower layers. It consists of $M-1$ blocks, each of which learns to map $\mathbf{F}_i^{m+1}$ back to $\mathbf{F}_i^{m}$. The block is stacked with a transpose convolutional layer, batch normalization layer and ReLU layer. We denote the feature maps of each block as $\{\mathbf{F'}_i^m\}_{m=1}^{M-1}$ in a top-down order and their attention maps $\{\mathbf{A'}_i^m\}_{m=1}^{M-1}$ are calculated by Eq.\,(\ref{equation:attention_calculation}). Each student is equipped with a self-distillation module, since each student has its own feature mapping. We train the self-distillation module together with the students, and in turn use it to distill the fused/diversity-enhanced features to the student model.

\begin{figure*}[!t]
\begin{center}
\includegraphics[width=0.9\linewidth, height=0.5\columnwidth]{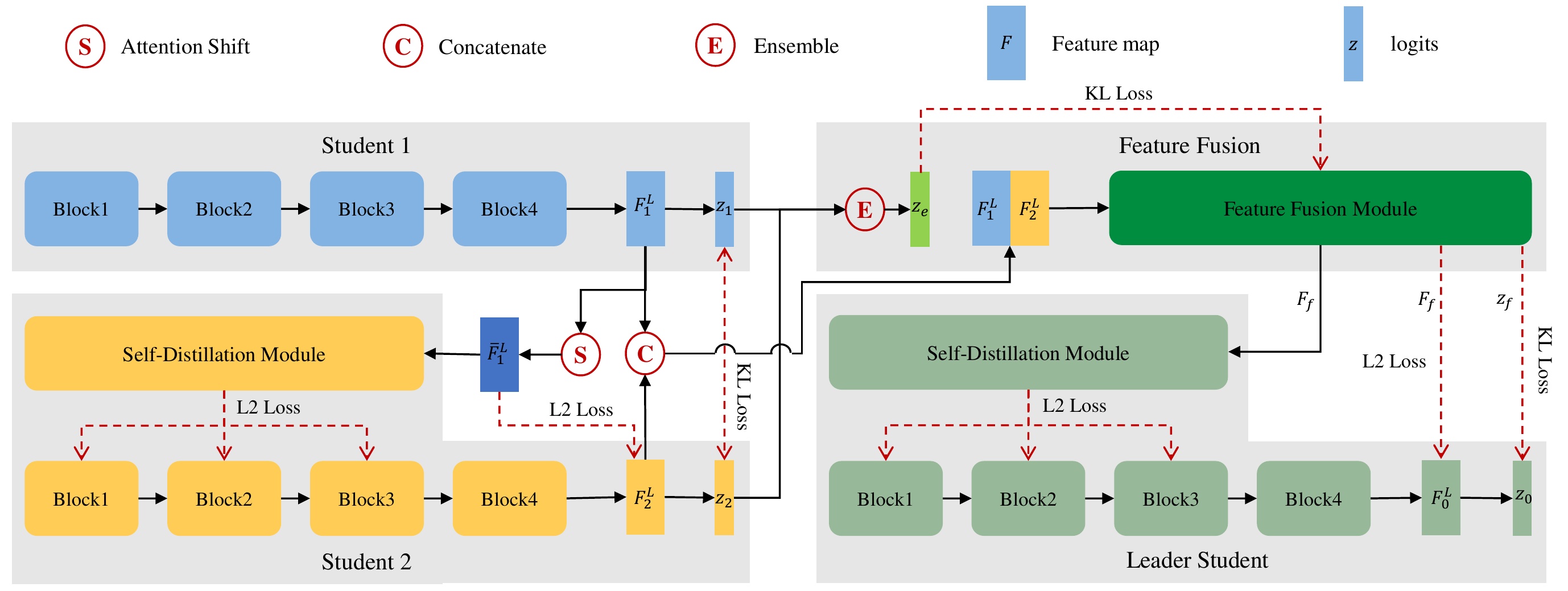}
\end{center}
\vspace{-1.0em}
\caption{Framework of our FFSD. First, student 1 and student 2 learn from each other in a collaborative way. Then, by shifting the attention of student 1 and distilling it to student 2, the diversity is enhanced among students. Lastly, the feature fusion module fuses all the students' information into a fused feature map. The fused representation is then used to assist the learning of the leader student. After training, we simply adopt the leader student for deployment.}
\vspace{-1.0em}
\label{fig:framework}
\end{figure*}

\textbf{Training of Self-Distillation Module.} As shown in Fig\,\ref{fig:self_distillation_training}, the self-distillation module takes the feature maps $\{\mathbf{F}_i^m\}_{m=1}^M$ as the training datasets, in which the feature map $\mathbf{F}_i^M$ of the last-layer output serves as the input, and the feature maps $\{\mathbf{F}_i^m\}_{m=1}^{M-1}$ serve as the target of each block. For student $\mathbf{S}_i$, the training objective of its self-distillation module is
\begin{equation}
\label{equation:self_distillation_module_loss}
\resizebox{\linewidth}{!}{$
\begin{aligned}
    \mathcal{L}_{\text{sdm}}=\sum_{m=1}^{M-1}\bigg\|\dfrac{\mathbf{A'}_i^m}{\|\mathbf{A'}_i^m\|_2} - \dfrac{\mathbf{A}_i^{m}}{\|\mathbf{A}_i^{m}\|_2}\bigg\|_2+
    \alpha\sum_{m=1}^{M-1}\bigg\|\dfrac{\mathbf{F'}_i^m}{\|\mathbf{F'}_i^m\|_2} - \dfrac{\mathbf{F}_i^{m}}{\|\mathbf{F}_i^{m}\|_2}\bigg\|_2,
\end{aligned}
$}
\end{equation}
where $\alpha$ balances the two loss terms. Due to the limited learning ability of the self-distillation module, we prefer to use it to learn the simple mapping of attention maps. Though an individual self-distillation block cannot completely map $\mathbf{F}_i^{m+1}$ back to $\mathbf{F}_i^{m}$, we encourage its output $\mathbf{F'}_i^{m}$ to be close to $\mathbf{F}_i^{m}$.

\begin{algorithm}[!t]
\begin{algorithmic}[1]
\caption{\label{alg1}Online knowledge distillation using Feature Fusion and Self-Distillation (FFSD)}
\REQUIRE A common student set $\{\textbf{S}_i\}_{i=1}^n$, a leader student $\textbf{S}_0$, a feature fusion module $\textbf{FF}$ and self-distillation modules $\{\textbf{SD}\}_{i=0}^n$.
\ENSURE A powerful leader student $\textbf{S}_0$.

\FOR {iter = 1 : Iter}
    \FOR{i = 1 : n}
        \STATE Forward the common student $\mathbf{S}_i$ and compute intermediate feature maps $\{\mathbf{F}_i^m\}_{m=1}^M$.
    \ENDFOR
    \STATE Forward the leader student $\mathbf{S}_0$.
    \STATE Forward the feature fusion module $\textbf{FF}$ using the feature map $\mathbf{F}_e$ concatenated by $\{\mathbf{F}_i^M\}_{i=1}^n$.
    \STATE Update the common student $\mathbf{S}_1$ using Eq.\,(\ref{equation:baseloss}).
    \FOR{i = 2 : n}
        \STATE Update the common student $\mathbf{S}_i$ using Eq.\,(\ref{equation:common_student_loss}).
    \ENDFOR
    \STATE Update the leader student $\mathbf{S}_0$ using Eq.\,(\ref{equation:student_leader_loss}).
    \STATE Update the feature fusion module $\textbf{FF}$ using Eq.\,(\ref{equation:fusion_module_loss}).
    \FOR{i = 0 : n}
        \STATE Update the self-distillation module $\mathbf{SD}_i$ using Eq.\,(\ref{equation:self_distillation_module_loss}).
    \ENDFOR
\ENDFOR
\end{algorithmic}
\end{algorithm}

% \begin{table*}[!t]
% \centering
% \caption{Experimental results on CIFAR-100. The ``Baseline'' trains the model using ground-truth labels only, ``2Net Avg'' represents the average accuracy of the two common students, and ``Ens'' represents the accuracy of the correct prediction of either student network. ``Fusion'' and ``Leader'' represent the accuracy of the fusion classifier and the leader student, respectively. All results reported are computed as the mean (standard deviations) of three runs.}
% \label{table:cifar100_result}
% \begin{tabular}{c|ccccccc}
% \toprule
% Model          & Baseline (\%) & 2Net Avg (\%) & Ens (\%) & Fusion (\%) & Leader (\%) & Gain($\uparrow$)        \\ \midrule
% ResNet-20      & 68.58 ± 0.26 & 71.92 ± 0.19 & 77.44 ± 0.14 & 73.43 ± 0.13 & 72.70 ± 0.13 & 4.12 ± 0.35 \\
% ResNet-32      & 69.96 ± 0.30 & 74.25 ± 0.08 & 79.93 ± 0.11 & 76.04 ± 0.17 & 74.85 ± 0.05 & 4.90 ± 0.25 \\
% ResNet-56      & 71.55 ± 0.50 & 75.61 ± 0.32 & 81.51 ± 0.16 & 77.28 ± 0.26 & 75.80 ± 0.11 & 4.25 ± 0.48 \\
% WRN-16-2       & 71.97 ± 0.09 & 75.41 ± 0.21 & 80.26 ± 0.25 & 76.69 ± 0.11 & 75.81 ± 0.08 & 3.83 ± 0.03 \\
% WRN-40-2       & 75.58 ± 0.17 & 78.73 ± 0.22 & 83.95 ± 0.26 & 80.24 ± 0.29 & 79.14 ± 0.03 & 3.47 ± 0.07 \\
% GoogLeNet      & 78.28 ± 0.24 & 81.48 ± 0.10 & 85.28 ± 0.13 & 82.42 ± 0.08 & 81.60 ± 0.25 & 3.32 ± 0.43 \\
% DenseNet-40-12 & 73.70 ± 0.10 & 76.87 ± 0.14 & 81.60 ± 0.12 & 78.34 ± 0.14 & 77.39 ± 0.23 & 3.68 ± 0.33 \\ \bottomrule
% \end{tabular}
% \end{table*}

\textbf{Application of Self-Distillation Module.} Thanks to the good feature mapping ability of the self-distillation module, we can distill the last layer of the diversity enhancement objective to the shallower layers through the self-distillation module as shown in Fig.\,\ref{fig:self_distillation_inference}. This achieves the diversity enhancement of the whole network, while ensuring the attention consistency and task dependence mentioned in Section~\ref{FeatureFusion}. Besides, we can use the self-distillation to distill the fused feature map to guide the training of shallower layers of the leader student.

For diversity enhancement, we first transform the diversity attention objective $\bar{\mathbf{A}}_{i-1}^M$ back into the diversity feature objective $\bar{\mathbf{F}}_{i-1}^M$ as the input of the self-distillation module. Then, the self-distillation module outputs the diversity objective $\{ {\bar{\mathbf{A}'}}_{i-1}^m\}_{m=1}^{M-1}$ of the shallower layers. The diversity enhancement objective of student $\mathbf{S}_i (i > 1)$ thus becomes
\begin{equation}
\label{equation:newdiversity_loss}
\resizebox{\linewidth}{!}{$
\mathcal{L}_{\text{div}}=\sum_{m=1}^{M-1}\bigg\|\dfrac{\mathbf{A}_i^m}{\|\mathbf{A}_i^m\|_2}-\dfrac{\bar{\mathbf{A}'}_{i-1}^{m}}{\|\bar{\mathbf{A}'}_{i-1}^{m}\|_2}\bigg\|_2 +  \bigg\|\dfrac{\mathbf{A}_i^M}{\|\mathbf{A}_i^M\|_2}-\dfrac{\bar{\mathbf{A}}_{i-1}^M}{\|\bar{\mathbf{A}}_{i-1}^M\|_2}\bigg\|_2.
$}
\end{equation}

The final objective of the common students is as follows:
\begin{equation}
\label{equation:common_student_loss}
    \mathcal{L}_{\text{stu}}=\mathcal{L}_{\text{base}}+\lambda_{\text{div}}\mathcal{L}_{\text{div}},
\end{equation}
where $\lambda_{\text{div}}$ controls the importance of each term.

Similarly, we use our proposed self-distillation module to distill the fused feature map to shallower layers of the leader student. Specifically, the self-distillation module of the leader student takes the fusion feature map $\mathbf{F}_f$ as its input, and outputs the shallower layers' target feature maps $\{\mathbf{F}^{*m}\}^{M-1}_{m=1}$. The corresponding self-distillation loss becomes
\begin{equation}
\label{equation:student_leader_distillation_loss}
    \mathcal{L}_{self}=\sum_{m=1}^{M-1}\bigg\|\dfrac{\mathbf{F}_0^m}{\|\mathbf{F}_0^m\|_2}-\dfrac{\mathbf{F}^{*m}}{\|\mathbf{F}^{*m}\|_2}\bigg\|_2.
\end{equation}

The final objective of the leader student is as follows:
\begin{equation}
\label{equation:student_leader_loss}
    \mathcal{L}_{\mathbf{S}_0}=\mathcal{L}_{CE}+T^2\mathcal{L}_{KL}+\lambda_{\text{fea}}\mathcal{L}_{\mathbf{F}_0^L}+\lambda_{\text{self}}\mathcal{L}_{\text{self}},
\end{equation}
where $\lambda_{\text{fea}}$ and $\lambda_{\text{self}}$ control the importance of each term. 

The overall framework of our FFSD is illustrated in Fig.\,\ref{fig:framework}. The training process can be referred to Alg.\,\ref{alg1}.

\section{Experiments}

\subsection{Experimental Settings}
\textbf{Datasets and Architecture.} To evaluate the efficacy of our FFSD online knowledge distillation method, we conduct experiments on two widely used datasets, CIFAR-100~\cite{krizhevsky2009learning} and ImageNet~\cite{russakovsky2015imagenet}. CIFAR-100 contains 50k images with 100 object classes for training and 10k images for testing. ImageNet is a large-scale dataset containing 1.28M training images and 50k validation images of 1000 object classes. The size of each image is 32$\times$32 for CIFAR-100 and 224$\times$224 for ImageNet. To verify the generalization of our proposed method on different network architectures, we conduct experiments using ResNet~\cite{he2016deep}, WRN~\cite{zagoruyko2016wide}, GoogLeNet~\cite{szegedy2015going}, DenseNet~\cite{huang2017densely}.

\textbf{Implementation Details.} We use stochastic gradient descent (SGD) with Nesterov momentum to optimize the training objective. The initial learning rate, momentum and weight decay are set to 0.1, 0.9 and 1e-4, respectively. For CIFAR-100, the models are trained with a batch size of 128 for 300 epochs and the learning rate is divided by 10 after 150 and 225 epochs. For ImageNet, we train all student models for 90 epochs with a batch size of 256. The learning rate warms up to 0.8 linearly in five epochs and is divided by 10 after 30 and 60 epochs. The feature fusion module uses the same implementation details described above. The self-distillation module is optimized by the ADAM optimizer with an initial learning rate of 0.001, using the same batch size, weight decay and learning rate decay strategy as the student model. The number of common students and temperature T are set to 2.

\begin{table*}[!t]
\centering
\caption{Experimental results on CIFAR-100. The ``Baseline'' trains the model using ground-truth labels only, and ``Ens'' represents the accuracy of the correct prediction of either student network. ``Fusion'' and ``Leader'' represent the accuracy of the fusion classifier and the leader student, respectively. All results are computed as the mean (standard deviations) of three runs.}
\label{table:cifar100_result}
\begin{tabular}{c|ccccccc}
\toprule
Model          & Baseline (\%) &Common Student1 (\%) & Common Student2 (\%) & Ens (\%) & Fusion (\%) & Leader (\%) & Gain($\uparrow$)        \\ \midrule
ResNet-20      & 68.58 ± 0.26 & 72.07 ± 0.21 & 71.95 ± 0.12 & 77.44 ± 0.14 & 73.43 ± 0.13 & 72.70 ± 0.13 & 4.12 ± 0.35 \\
ResNet-32      & 69.96 ± 0.30 &74.30 ± 0.10 &74.34 ± 0.21 & 79.93 ± 0.11 & 76.04 ± 0.17 & 74.85 ± 0.05 & 4.90 ± 0.25 \\
ResNet-56      & 71.55 ± 0.50 &75.64 ± 0.23 &75.78 ± 0.43 & 81.51 ± 0.16 & 77.28 ± 0.26 & 75.80 ± 0.11 & 4.25 ± 0.48 \\
WRN-16-2       & 71.97 ± 0.09 &75.51 ± 0.12 & 75.44 ± 0.25 & 80.26 ± 0.25 & 76.69 ± 0.11 & 75.81 ± 0.08 & 3.83 ± 0.03 \\
WRN-40-2       & 75.58 ± 0.17 &78.85 ± 0.26 & 78.83 ± 0.21 & 83.95 ± 0.26 & 80.24 ± 0.29 & 79.14 ± 0.03 & 3.47 ± 0.07 \\
GoogLeNet      & 78.28 ± 0.24 & 81.51 ± 0.07 & 81.58 ± 0.16 & 85.28 ± 0.13 & 82.42 ± 0.08 & 81.60 ± 0.25 & 3.32 ± 0.43 \\
DenseNet-40-12 & 73.70 ± 0.10 & 76.87 ± 0.25 & 77.01 ± 0.06 & 81.60 ± 0.12 & 78.34 ± 0.14 & 77.39 ± 0.23 & 3.68 ± 0.33 \\ \bottomrule
\end{tabular}
\end{table*}

Our hyper-parameters include $\lambda_{\text{div}}$, $\lambda_{\text{fea}}$, $\lambda_{\text{self}}$, and $\alpha$. For each hyper-parameter, we search its optimal value using grid search with others fixed. For different networks and datasets, this process can be used to find the corresponding optimal values. We conduct grid search for ResNet-32 on CIFAR-100 and obtain $\lambda_{\text{div}}$ = 1e-5, $\lambda_{\text{fea}}$ = 10, $\lambda_{\text{self}}$= 1e3 and $\alpha$ = 1. These settings are applied to other networks and datasets. Though not optimal, they already provide the state-of-the-art results.

\begin{table}[!t]
\caption{Results of our FFSD compared with several state-of-the-art methods on CIFAR-100. In KD and AT, we use the pre-trained ResNet-56 and WRN-40-2 as the teacher model of ResNet-32 and WRN-16-2, respectively.}
\label{table:cifar100_compare}
\resizebox{\columnwidth}{!}{
\begin{tabular}{c|cc|cc}
\toprule
Method     & \begin{tabular}[c]{@{}c@{}}ResNet-32\\ (\%)\end{tabular} & Gain($\uparrow$) & \begin{tabular}[c]{@{}c@{}}WRN-16-2\\ (\%)\end{tabular} & Gain($\uparrow$) \\ \midrule
Baseline   & 69.96     & -    & 71.97   & -    \\
KD~\cite{hinton2015distilling}         & 72.87     & 2.91 & 73.79   & 1.82 \\
AT~\cite{komodakis2017paying}         & 71.23     & 1.27 & 73.70   & 1.73 \\ \midrule
DML~\cite{zhang2018deep}        & 73.64     & 3.68 & 74.63   & 2.66 \\
AFD~\cite{chung2020feature}        & 74.03     & 4.07 & 75.33   & 3.36 \\
AMLN~\cite{zhang2020amln}       & 74.69     & 4.73 & 75.56   & 3.59 \\ \midrule
ONE~\cite{zhu2018knowledge}        & 73.39     & 3.43 & 74.84   & 2.87 \\
FFL~\cite{kim2019feature}        & 74.44     & 4.48 & 75.26   & 3.29 \\
KDCL~\cite{guo2020online}       & 74.30     & 4.34 & 75.50   & 3.53 \\ 
OKDDip~\cite{chen2020online}     & 74.60     & 4.64 & 75.31   & 3.34 \\ \midrule
\textbf{FFSD(Ours)} & \textbf{74.85}     & \textbf{4.90} & \textbf{75.81}   & \textbf{3.84} \\ \bottomrule
\end{tabular}
}
\end{table}

\subsection{Experimental Results}
\textbf{Results on CIFAR-100.} We first evaluate FFSD on CIFAR-100 in Tab.\,\ref{table:cifar100_result}. After feature fusion, the fusion classifier improves the accuracy over the baseline and students. This demonstrates that fusing the information of all students is of great help to the final result, and it is inappropriate to only adopt the optimal student. However, such improvements require retention of all students, increasing the storage and inference cost. Thus, we distill the knowledge from the feature fusion module to the leader student, which yields a 3.32\%-4.90\% improvement. Among the models compared, FFSD achieves a 4.90\% improvement on ResNet-32 and 3.83\% on WRN-16-2, results of which are superior to all other students.

\textbf{Quantitative Comparison on CIFAR-100.} As shown in Tab.\,\ref{table:cifar100_compare}, we compare FFSD with several state-of-the-art methods on ResNet-32 and WRN-16-2. FFSD surpasses most knowledge distillation methods, including traditional knowledge distillation KD~\cite{hinton2015distilling} and AT~\cite{komodakis2017paying}, mutual learning based DML~\cite{zhang2018deep}, AFD~\cite{chung2020feature} and AMLN~\cite{zhang2020amln}, and ensemble learning based ONE~\cite{zhu2018knowledge}, FFL~\cite{kim2019feature}, KDCL~\cite{guo2020online}, and OKDDip~\cite{chen2020online}. For example, with ResNet-32, FFSD achieves an accuracy of 74.85\%, which is higher than AMLN's 74.69\%. In addition, with WRN-16-2, FFSD can achieve a 3.84\% performance improvement, which is superior to AMLN's 3.59\% and KDCL's 3.53\%.
Similarly, EnD2~\cite{malinin2020ensemble} also distills the prediction distribution from an ensemble into a single model. Following EnD2~\cite{malinin2020ensemble}, we compare FFSD with EnD2 on VGG-16~\cite{simonyan2014very}. FFSD achieves an accuracy of 75.87\% while End2 has only 73.7\%. It is worth noting that EnD2 needs an ensemble of 10 models to capture the diversity of the ensemble, which makes EnD2 hard to be extended to a larger model, like ResNet-34. 
%It fully shows the advancement and superiority of our FFSD.

\textbf{Results on ImageNet.} We further conduct experiments on the large-scale ImageNet dataset. We choose two most popular models, ResNet-18 and ResNet-34, for verification. As shown in Tab.\,\ref{table:imagenet_compare}, with ResNet-18, FFSD achieves 70.9\% top-1 accuracy, which is superior to that of DML and ONE. Besides, Common student1 and common student2 achieve 70.15\% and 70.18\% accuracy respectively, which are also higher than baseline.
For ResNet-34, FFSD can obtain 1.5\% gains, outperforming the baseline model and other online distillation methods. 
Even the performance of common students (student1: 74.16\% \& student2: 74.25\%) in FFSD is higher than that of ONE.
Hence, FFSD can well generalize to complex datasets.

\begin{table}[!t]
\centering
\caption{Results of our FFSD compared with several online knowledge distillation methods on ImageNet.}
\label{table:imagenet_compare}
\begin{tabular}{c|ccc}
\toprule
\multicolumn{1}{c|}{Model} & Method    & Top1-Acc (\%) & Gain($\uparrow$) \\ \midrule
\multirow{5}{*}{ResNet-18} & Baseline   & 69.7     & -    \\
                           & DML~\cite{zhang2018deep}        & 69.8     & 0.1  \\
                           & ONE~\cite{zhu2018knowledge}        & 70.2     & 0.5    \\
                           & KDCL~\cite{guo2020online} & 70.4 & 0.7  \\
                           & \textbf{FFSD(Ours)} &  \textbf{70.9}  &  \textbf{1.2} \\ \midrule
\multirow{5}{*}{ResNet-34} & Baseline   & 73.2    & -    \\
                           & DML~\cite{zhang2018deep}        & 74.0    & 0.8 \\
                           & ONE~\cite{zhu2018knowledge}        & 74.1    & 0.9 \\
                           & KDCL~\cite{guo2020online} & 74.4 & 1.2  \\
                           & \textbf{FFSD(Ours)} &     \textbf{74.7}    &   \textbf{1.5} \\ \bottomrule
\end{tabular}
\end{table}

\begin{table*}[!t]
\renewcommand\arraystretch{1.1}
\centering
\caption{Comparison between different diversity enhancement strategies. DML does not adopt any diversity enhancement strategy. L2 uses Eq.\,(\ref{equation:loss_dir}) for diversity enhancement. FFSD is our proposed diversity enhancement strategy. ``Cosine'' represents the cosine similarity between the two student models.}
\begin{tabular}{c|ccc|ccc|ccc}
\hline
\multirow{2}{*}{Model} & \multicolumn{3}{c|}{DML~\cite{zhang2018deep}}                          & \multicolumn{3}{c|}{L2}                                              & \multicolumn{3}{c}{FFSD}                                             \\ \cline{2-10} 
          & \multicolumn{1}{c|}{2Net Avg} & \multicolumn{1}{c|}{Fusion} & Cosine & \multicolumn{1}{c|}{2Net Avg} & \multicolumn{1}{c|}{Fusion} & Cosine & \multicolumn{1}{c|}{2Net Avg} & \multicolumn{1}{c|}{Fusion} & Cosine \\ \hline
ResNet-32 & 73.64\% & 75.15\% & 0.2635 & 73.19\% & 75.03\% & 0.2351 & 74.25\% & 76.04\% & 0.2550 \\
WRN-16-2  & 74.63\% & 76.18\% & 0.2787 & 74.69\% & 75.99\% & 0.2570 & 75.41\% & 76.69\% & 0.2491  \\ \hline
\end{tabular}
\label{table:necessity_diversity}
\end{table*}

\begin{figure}[!t]
\begin{center}
\includegraphics[width=0.75\linewidth]{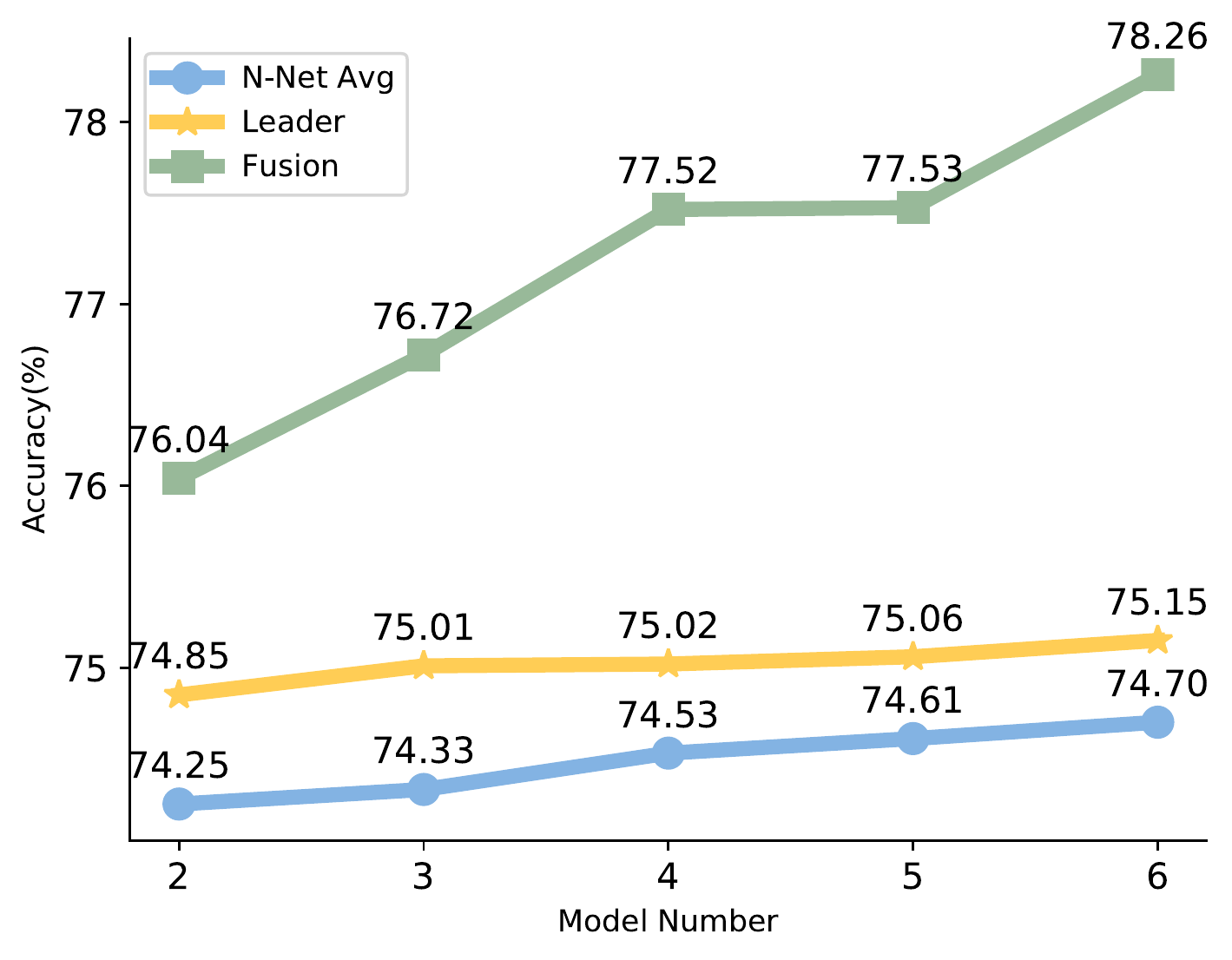}
\end{center}
\vspace{-1.5em}
\caption{Experimental results when increasing the number of students using ResNet-32 on CIFAR-100.}
\vspace{-1.2em}
\label{fig:mutil_models}
\end{figure}

\begin{table}[!t]
\centering
\caption{Experimental results of other online distillation methods combined with the leader student strategy on CIFAR-100. ``SL'' represents the leader student.}
\label{table:combine_student_leader_results}
\begin{tabular}{c|cc|cc}
\toprule
\multirow{2}{*}{Model} & \multicolumn{2}{c|}{DML~\cite{zhang2018deep}+SL}            & \multicolumn{2}{c}{ONE~\cite{zhu2018knowledge}+SL}           \\ \cline{2-5} 
                       & \multicolumn{1}{c|}{Original} & Leader & \multicolumn{1}{c|}{Original} & Leader  \\ \midrule
ResNet-32 & 73.64\% & 74.02\% & 73.39\% & 73.98\% \\ 
WRN-16-2  & 74.63\% & 75.15\% & 74.84\% & 75.44\%       \\ \bottomrule
\end{tabular}
\end{table}

\textbf{Impact of Student Number.} We increase the number of common students in Fig.\,\ref{fig:mutil_models}. The accuracy of the students, the leader student, and the fusion classifier increase as the number of students grows. In fact, the fusion classifier achieves an astonishing accuracy of 78.26\% after fusing the output feature maps of six students. Keeping only the optimal student after online knowledge distillation wastes the effective knowledge of other students. We observe that the accuracy of the leader student is always higher than the average accuracy of students, even when the number of students is up to six. %the average accuracy of the students is still lower than that of the leader student who only distilled by the two students. %While simply stacking the number of students can further improve the leader student, the computational overhead increment is far greater than the performance improvement.

\begin{table*}[!t]
\centering
\caption{The effect of the self-distillation module on CIFAR-100. L2 uses Eq.\,(\ref{equation:loss_dir}) for diversity enhancement, which causes attention inconsistency. ``SD'' represents our proposed self-distillation module.}
\label{table:self_distillation_effects}
\resizebox{\linewidth}{!}{
\begin{tabular}{c|ccc|ccc|ccc|ccc}
\toprule
\multirow{2}{*}{Model} & \multicolumn{3}{c|}{L2}                                              & \multicolumn{3}{c|}{L2 with SD}                                      & \multicolumn{3}{c|}{FFSD w/o SD}                                     & \multicolumn{3}{c}{FFSD}                                             \\ \cline{2-13} 
         & \multicolumn{1}{c|}{2Net Avg} & \multicolumn{1}{c|}{Fusion} & Leader & \multicolumn{1}{c|}{2Net Avg} & \multicolumn{1}{c|}{Fusion} & Leader & \multicolumn{1}{c|}{2Net Avg} & \multicolumn{1}{c|}{Fusion} & Leader & \multicolumn{1}{c|}{2Net Avg} & \multicolumn{1}{c|}{Fusion} & Leader \\ \midrule
ResNet-32 & 73.19\% & 75.03\% & 73.78\% & 73.75\% & 75.48\% & 74.27\% & 73.69\% & 75.61\% & 73.94\% & 74.25\% & 76.04\% & 74.85\%  \\
WRN-16-2 & 74.69\% & 75.99\% & 75.20\% & 75.00\% & 76.15\% & 75.33\% & 74.63\% & 76.05\% & 75.43\% & 75.41\% & 76.69\% & 75.81\%  \\ \bottomrule
\end{tabular}
}
\end{table*}

\subsection{Detailed Analysis}

\textbf{Necessity of the Diversity Enhancement Strategy.} 
We examine the effect of enhancing the diversity among students in the fusion classifier. As shown in Tab.\,\ref{table:necessity_diversity}, with ResNet-32, the cosine similarity of the two students in DML reaches 0.2635 without any diversity enhancement, and the fusion classifier can only achieve a 1.51\% improvement in accuracy. Although the L2 reduces the cosine similarity between the two students, minimizing Eq.\,(\ref{equation:loss_dir}) greatly affects the performance of the students. The goal of the $\ell_2$-norm loss is to minimize the overall average outputs, which lacks a clear objective. Our FFSD presents a clear diversity enhancement strategy for attention shifting, which decreases the cosine similarity between the two students and improves the performance of the models. Similar results are also observed on WRN-16-2, while the L2 slightly improves the performance of the two students, the performance of the fusion classifier declines.% Experimental results show that the proposed diversity enhancement strategy can effectively improve the performance of the fusion classifier.

\textbf{Advancement of the leader student and Feature Fusion.} We combine the learning strategy of the leader student in FFSD with other online knowledge distillation methods for futher experiments. From Tab.\,\ref{table:combine_student_leader_results}, we can see that the addition of the leader student injects vitality into other online knowledge distillation methods, and the performance of the leader student is improved. %Among them, with WRN-16-2, the combination of DML and the leader student reaches an accuracy of 75.15\%, and improves the performance by 0.52\%. 
Among them, the combination of DML and the leader student yields a 0.38\%-0.52\% improvement in accuracy over the original results. The leader student also improves the performance of ONE by 0.59\%-0.60\%. 
When removing feature fusion from FFSD, the performance significantly decreases from 74.85\% to 74.06\%. This well demonstrates the effectiveness of our feature fusion in helping student leader learn more information from common students. 
%
%The experimental results show the necessity of the leader student and the effectiveness of its learning strategy.

\begin{figure}[!t]
\begin{center}
\includegraphics[width=0.7\linewidth]{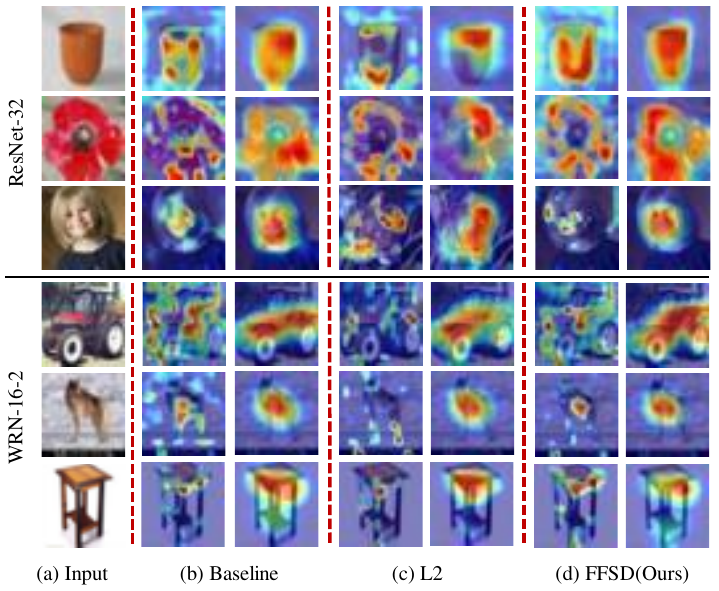}
\end{center}
\vspace{-1.2em}
\caption{Feature map visualization with and without self-distillation on ResNet-32 and WRN-16-2. For each method, the left and right parts show the mid-, top-level activation feature maps, respectively. }
\vspace{-1.2em}
\label{fig:self_distillation_effects}
\end{figure}

\textbf{Importance of the Self-Distillation.} In Fig.\,\ref{fig:self_distillation_effects}, we visualize the feature maps with and without self-distillation. There is a significant overlap of attention positions between the activation feature maps of the mid- and top-level of the baseline, which demonstrates the attention consistency. However, it is difficult for us to see this phenomenon with the L2. The activation feature maps of the mid- and top-level only overlap in a very small part. Note that, in the picture of the wolf in row 5 of the L2, there is no way to find the attention consistency phenomenon, which is one of the reasons for the poor performance. Even though FFSD also enhances diversity, our self-distillation addresses the problem of attention inconsistency.

Tab.\,\ref{table:self_distillation_effects} demonstrates the effect of self-distillation on the performance of the student models. We only retain the last-layer diversity enhancement loss calculation of the L2 to enhance diversity and add self-distillation to it. All indicators are improved. We also investigate the effect of removing the self-distillation from FFSD. Though the performance of students degrades, but it is still better than that with the L2. 
We also explore the cases that the self-distillation is only applied to common students or leader student. We respectively obtain accuracy of 74.71\% (common students) and 74.04\% (leader student). In contrast, our FFSD achieves the best performance of 74.85\% when the self-distillation module is applied to both common students and student leader.
%
%In conclusion, our self-distillation not only solves the problem of attention inconsistency during diversity enhancement, but also improves the performance of the student models.

\textbf{Training Time Consumption.} We measure the training time of WRN-16-1 on CIFAR-100. We use AT~\cite{komodakis2017paying} for comparison and select WRN-40-2 as its teacher model.
In Tab.~\ref{table:time_consumption}, we report the FLOPs of different components in our FFSD and AT as well as the practial training time. Though FFSD introduces additional operations from feature fusion and self distillation, it still achieves lower overhead and better performance (75.81\% \emph{vs}. 73.71\% by AT in Tab.\,\ref{table:cifar100_compare})

\begin{table}[]
\centering
\caption{Training time consumption of our FFSD. Our experiments are conducted on an Intel Xeno CPU E5-2690 CPU and one NVIDIA GTX1080TI GPU under Pytorch V1.5.1.}
\label{table:time_consumption}
\resizebox{\columnwidth}{!}{
\begin{tabular}{c|ccccc}
\toprule
Method                & Component                & \tabincell{c}{FLOPs\\(M)} & \tabincell{c}{Component\\Num} & \tabincell{c}{Total FLOPs\\(M)}          & \tabincell{c}{Training Time\\(Hour)} \\ \midrule
\multirow{3}{*}{FFSD} & WRN-16-2                 & 130.5     & 3                                                             & \multirow{3}{*}{420.15} & \multirow{3}{*}{4.4} \\
                      & Self-distillation & 14.3      & 2                                                             &                         &                      \\
                      & Feature Fusion    & 0.05      & 1                                                             &                         &                      \\ \midrule
\multirow{2}{*}{AT~\cite{komodakis2017paying}}   & WRN-40-2                 & 358.3     & 1                                                             & \multirow{2}{*}{488.80} & \multirow{2}{*}{4.6} \\
                      & WRN-16-2                 & 130.5     & 1                                                             &                         &               \\ \bottomrule
\end{tabular}
}
\end{table}

\section{Conclusion}
In this paper, a novel online knowledge distillation method, termed FFSD, is proposed  using feature fusion and self-distillation. Existing online knowledge distillation methods either adopt the student with the best performance or consider the holistic performance using an ensemble model. However, they either ignore other students' information or increase the computational burden during deployment. To solve these issues, we first design a feature fusion module similar to an autoencoder to fuse the output feature maps from all students into a meaningful and compact fused feature map, which is then distilled to the leader student. At the same time, we design a diversity enhancement strategy to enhance the diversity among students, enabling the leader student to obtain more information during feature fusion. Second, a self-distillation module is proposed to convert the feature maps of deeper layers to shallower ones, which are then distilled to shallower layers. This increases the generalization ability of the model. Extensive experiments on CIFAR-100 and ImageNet demonstrate the superiority of our FFSD.

% \fi
\IEEEpeerreviewmaketitle

\bibliographystyle{IEEEtran}
\bibliography{reference}

\begin{thebibliography}{10}
\providecommand{\url}[1]{#1}
\csname url@samestyle\endcsname
\providecommand{\newblock}{\relax}
\providecommand{\bibinfo}[2]{#2}
\providecommand{\BIBentrySTDinterwordspacing}{\spaceskip=0pt\relax}
\providecommand{\BIBentryALTinterwordstretchfactor}{4}
\providecommand{\BIBentryALTinterwordspacing}{\spaceskip=\fontdimen2\font plus
\BIBentryALTinterwordstretchfactor\fontdimen3\font minus
  \fontdimen4\font\relax}
\providecommand{\BIBforeignlanguage}[2]{{%
\expandafter\ifx\csname l@#1\endcsname\relax
\typeout{** WARNING: IEEEtran.bst: No hyphenation pattern has been}%
\typeout{** loaded for the language `#1'. Using the pattern for}%
\typeout{** the default language instead.}%
\else
\language=\csname l@#1\endcsname
\fi
#2}}
\providecommand{\BIBdecl}{\relax}
\BIBdecl

\bibitem{han2015learning}
S.~Han, J.~Pool, J.~Tran, and W.~Dally, ``Learning both weights and connections
  for efficient neural network,'' in \emph{Proceedings of the Advances in
  Neural Information Processing Systems (NeurIPS)}, 2015, pp. 1135--1143.

\bibitem{he2019filter}
Y.~He, P.~Liu, Z.~Wang, Z.~Hu, and Y.~Yang, ``Filter pruning via geometric
  median for deep convolutional neural networks acceleration,'' in
  \emph{Proceedings of the IEEE Conference on Computer Vision and Pattern
  Recognition (CVPR)}, 2019, pp. 4340--4349.

\bibitem{rastegari2016xnor}
M.~Rastegari, V.~Ordonez, J.~Redmon, and A.~Farhadi, ``Xnor-net: Imagenet
  classification using binary convolutional neural networks,'' in
  \emph{Proceedings of the European Conference on Computer Vision
  (ECCV)}.\hskip 1em plus 0.5em minus 0.4em\relax Springer, 2016, pp. 525--542.

\bibitem{jacob2018quantization}
B.~Jacob, S.~Kligys, B.~Chen, M.~Zhu, M.~Tang, A.~Howard, H.~Adam, and
  D.~Kalenichenko, ``Quantization and training of neural networks for efficient
  integer-arithmetic-only inference,'' in \emph{Proceedings of the IEEE
  Conference on Computer Vision and Pattern Recognition (CVPR)}, 2018, pp.
  2704--2713.

\bibitem{denton2014exploiting}
E.~L. Denton, W.~Zaremba, J.~Bruna, Y.~LeCun, and R.~Fergus, ``Exploiting
  linear structure within convolutional networks for efficient evaluation,'' in
  \emph{Proceedings of the Advances in Neural Information Processing Systems
  (NeurIPS)}, 2014, pp. 1269--1277.

\bibitem{zhang2015efficient}
X.~Zhang, J.~Zou, X.~Ming, K.~He, and J.~Sun, ``Efficient and accurate
  approximations of nonlinear convolutional networks,'' in \emph{Proceedings of
  the IEEE Conference on Computer Vision and Pattern Recognition (CVPR)}, 2015,
  pp. 1984--1992.

\bibitem{hinton2015distilling}
G.~Hinton, O.~Vinyals, and J.~Dean, ``Distilling the knowledge in a neural
  network,'' \emph{arXiv preprint arXiv:1503.02531}, 2015.

\bibitem{romero2014fitnets}
A.~Romero, N.~Ballas, S.~E. Kahou, A.~Chassang, C.~Gatta, and Y.~Bengio,
  ``Fitnets: Hints for thin deep nets,'' \emph{arXiv preprint arXiv:1412.6550},
  2014.

\bibitem{komodakis2017paying}
N.~Komodakis and S.~Zagoruyko, ``Paying more attention to attention: improving
  the performance of convolutional neural networks via attention transfer,'' in
  \emph{Proceedings of the International Conference of Learning Representation
  (ICLR)}, 2017.

\bibitem{zhang2018deep}
Y.~Zhang, T.~Xiang, T.~M. Hospedales, and H.~Lu, ``Deep mutual learning,'' in
  \emph{Proceedings of the IEEE Conference on Computer Vision and Pattern
  Recognition (CVPR)}, 2018, pp. 4320--4328.

\bibitem{song2018collaborative}
G.~Song and W.~Chai, ``Collaborative learning for deep neural networks,'' in
  \emph{Proceedings of the Advances in Neural Information Processing Systems
  (NeurIPS)}, 2018, pp. 1832--1841.

\bibitem{zhu2018knowledge}
X.~Zhu, S.~Gong \emph{et~al.}, ``Knowledge distillation by on-the-fly native
  ensemble,'' in \emph{Proceedings of the Advances in Neural Information
  Processing Systems (NeurIPS)}, 2018, pp. 7517--7527.

\bibitem{jin2019knowledge}
X.~Jin, B.~Peng, Y.~Wu, Y.~Liu, J.~Liu, D.~Liang, J.~Yan, and X.~Hu,
  ``Knowledge distillation via route constrained optimization,'' in
  \emph{Proceedings of the IEEE International Conference on Computer Vision
  (ICCV)}, 2019, pp. 1345--1354.

\bibitem{cho2019efficacy}
J.~H. Cho and B.~Hariharan, ``On the efficacy of knowledge distillation,'' in
  \emph{Proceedings of the IEEE International Conference on Computer Vision
  (ICCV)}, 2019, pp. 4794--4802.

\bibitem{chung2020feature}
I.~Chung, S.~Park, J.~Kim, and N.~Kwak, ``Feature-map-level online adversarial
  knowledge distillation,'' in \emph{Proceedings of the International
  Conference on Machine Learning (ICML)}, 2020, pp. 2006--2015.

\bibitem{zhang2020amln}
X.~Zhang, S.~Lu, H.~Gong, Z.~Luo, and M.~Liu, ``Amln: Adversarial-based mutual
  learning network for online knowledge distillation,'' in \emph{Proceedings of
  the European Conference on Computer Vision (ECCV)}, 2020, pp. 158--173.

\bibitem{chen2020online}
D.~Chen, J.-P. Mei, C.~Wang, Y.~Feng, and C.~Chen, ``Online knowledge
  distillation with diverse peers.'' in \emph{Proceedings of the AAAI
  Conference on Artificial Intelligence (AAAI)}, 2020, pp. 3430--3437.

\bibitem{guo2020online}
Q.~Guo, X.~Wang, Y.~Wu, Z.~Yu, D.~Liang, X.~Hu, and P.~Luo, ``Online knowledge
  distillation via collaborative learning,'' in \emph{Proceedings of the IEEE
  Conference on Computer Vision and Pattern Recognition (CVPR)}, 2020, pp.
  11\,020--11\,029.

\bibitem{kim2019feature}
J.~Kim, M.~Hyun, I.~Chung, and N.~Kwak, ``Feature fusion for online mutual
  knowledge distillation,'' \emph{arXiv preprint arXiv:1904.09058}, 2019.

\bibitem{yuan2020revisiting}
L.~Yuan, F.~E. Tay, G.~Li, T.~Wang, and J.~Feng, ``Revisiting knowledge
  distillation via label smoothing regularization,'' in \emph{Proceedings of
  the IEEE Conference on Computer Vision and Pattern Recognition (CVPR)}, 2020,
  pp. 3903--3911.

\bibitem{ding2021knowledge}
Q.~Ding, S.~Wu, T.~Dai, H.~Sun, J.~Guo, Z.-H. Fu, and S.~Xia, ``Knowledge
  refinery: Learning from decoupled label,'' in \emph{Proceedings of the AAAI
  Conference on Artificial Intelligence (AAAI)}, 2021, pp. 7228--7235.

\bibitem{chen2021distilling}
P.~Chen, S.~Liu, H.~Zhao, and J.~Jia, ``Distilling knowledge via knowledge
  review,'' in \emph{Proceedings of the IEEE Conference on Computer Vision and
  Pattern Recognition (CVPR)}, 2021, pp. 5008--5017.

\bibitem{chen2021cross}
D.~Chen, J.-P. Mei, Y.~Zhang, C.~Wang, Z.~Wang, Y.~Feng, and C.~Chen,
  ``Cross-layer distillation with semantic calibration,'' in \emph{Proceedings
  of the AAAI Conference on Artificial Intelligence (AAAI)}, 2021, pp.
  7028--7036.

\bibitem{yim2017gift}
J.~Yim, D.~Joo, J.~Bae, and J.~Kim, ``A gift from knowledge distillation: Fast
  optimization, network minimization and transfer learning,'' in
  \emph{Proceedings of the IEEE Conference on Computer Vision and Pattern
  Recognition (CVPR)}, 2017, pp. 4133--4141.

\bibitem{tian2019contrastive}
Y.~Tian, D.~Krishnan, and P.~Isola, ``Contrastive representation
  distillation,'' \emph{arXiv preprint arXiv:1910.10699}, 2019.

\bibitem{xu2020knowledge}
G.~Xu, Z.~Liu, X.~Li, and C.~C. Loy, ``Knowledge distillation meets
  self-supervision,'' \emph{arXiv preprint arXiv:2006.07114}, 2020.

\bibitem{park2019relational}
W.~Park, D.~Kim, Y.~Lu, and M.~Cho, ``Relational knowledge distillation,'' in
  \emph{Proceedings of the IEEE Conference on Computer Vision and Pattern
  Recognition (CVPR)}, 2019, pp. 3967--3976.

\bibitem{liu2019knowledge}
Y.~Liu, J.~Cao, B.~Li, C.~Yuan, W.~Hu, Y.~Li, and Y.~Duan, ``Knowledge
  distillation via instance relationship graph,'' in \emph{Proceedings of the
  IEEE Conference on Computer Vision and Pattern Recognition (CVPR)}, 2019, pp.
  7096--7104.

\bibitem{passalis2020heterogeneous}
N.~Passalis, M.~Tzelepi, and A.~Tefas, ``Heterogeneous knowledge distillation
  using information flow modeling,'' in \emph{Proceedings of the IEEE
  Conference on Computer Vision and Pattern Recognition (CVPR)}, 2020, pp.
  2339--2348.

\bibitem{li2017learning}
Y.~Li, J.~Yang, Y.~Song, L.~Cao, J.~Luo, and L.-J. Li, ``Learning from noisy
  labels with distillation,'' in \emph{Proceedings of the IEEE International
  Conference on Computer Vision (ICCV)}, 2017, pp. 1910--1918.

\bibitem{wang2018kdgan}
X.~Wang, R.~Zhang, Y.~Sun, and J.~Qi, ``Kdgan: Knowledge distillation with
  generative adversarial networks,'' in \emph{Proceedings of the Advances in
  Neural Information Processing Systems (NeurIPS)}, 2018, pp. 775--786.

\bibitem{goodfellow2014generative}
I.~Goodfellow, J.~Pouget-Abadie, M.~Mirza, B.~Xu, D.~Warde-Farley, S.~Ozair,
  A.~Courville, and Y.~Bengio, ``Generative adversarial nets,'' in
  \emph{Proceedings of the Advances in Neural Information Processing Systems
  (NeurIPS)}, 2014, pp. 2672--2680.

\bibitem{jang2019learning}
Y.~Jang, H.~Lee, S.~J. Hwang, and J.~Shin, ``Learning what and where to
  transfer,'' in \emph{Proceedings of the International Conference on Machine
  Learning (ICML)}, 2019, pp. 3030--3039.

\bibitem{liu2020search}
Y.~Liu, X.~Jia, M.~Tan, R.~Vemulapalli, Y.~Zhu, B.~Green, and X.~Wang, ``Search
  to distill: Pearls are everywhere but not the eyes,'' in \emph{Proceedings of
  the IEEE Conference on Computer Vision and Pattern Recognition (CVPR)}, 2020,
  pp. 7539--7548.

\bibitem{li2020gan}
M.~Li, J.~Lin, Y.~Ding, Z.~Liu, J.-Y. Zhu, and S.~Han, ``Gan compression:
  Efficient architectures for interactive conditional gans,'' in
  \emph{Proceedings of the IEEE Conference on Computer Vision and Pattern
  Recognition (CVPR)}, 2020, pp. 5284--5294.

\bibitem{li2020learning}
S.~Li, M.~Lin, Y.~Wang, F.~Chao, X.~Mao, M.~Xu, Y.~Wu, F.~Huang, L.~Shao, and
  R.~Ji, ``Learning efficient gans for image translation via differentiable
  masks and co-attention distillation,'' \emph{arXiv preprint
  arXiv:2011.08382}, 2020.

\bibitem{jin2021teachers}
Q.~Jin, J.~Ren, O.~J. Woodford, J.~Wang, G.~Yuan, Y.~Wang, and S.~Tulyakov,
  ``Teachers do more than teach: Compressing image-to-image models,'' in
  \emph{Proceedings of the IEEE Conference on Computer Vision and Pattern
  Recognition (CVPR)}, 2021, pp. 13\,600--13\,611.

\bibitem{liu2021content}
Y.~Liu, Z.~Shu, Y.~Li, Z.~Lin, F.~Perazzi, and S.-Y. Kung, ``Content-aware gan
  compression,'' in \emph{Proceedings of the IEEE Conference on Computer Vision
  and Pattern Recognition (CVPR)}, 2021, pp. 12\,156--12\,166.

\bibitem{gou2021knowledge}
J.~Gou, B.~Yu, S.~J. Maybank, and D.~Tao, ``Knowledge distillation: A survey,''
  \emph{International Journal of Computer Vision (IJCV)}, pp. 1789--1819, 2021.

\bibitem{wang2021knowledge}
L.~Wang and K.-J. Yoon, ``Knowledge distillation and student-teacher learning
  for visual intelligence: A review and new outlooks,'' \emph{IEEE Transactions
  on Pattern Analysis and Machine Intelligence (TPAMI)}, 2021.

\bibitem{malinin2020ensemble}
A.~Malinin, B.~Mlodozeniec, and M.~Gales, ``Ensemble distribution
  distillation,'' in \emph{Proceedings of the International Conference of
  Learning Representation (ICLR)}, 2020.

\bibitem{furlanello2018born}
T.~Furlanello, Z.~C. Lipton, M.~Tschannen, L.~Itti, and A.~Anandkumar, ``Born
  again neural networks,'' \emph{arXiv preprint arXiv:1805.04770}, 2018.

\bibitem{lee2019rethinking}
H.~Lee, S.~J. Hwang, and J.~Shin, ``Rethinking data augmentation:
  Self-supervision and self-distillation,'' \emph{arXiv preprint
  arXiv:1910.05872}, 2019.

\bibitem{xu2019data}
T.-B. Xu and C.-L. Liu, ``Data-distortion guided self-distillation for deep
  neural networks,'' in \emph{Proceedings of the AAAI Conference on Artificial
  Intelligence (AAAI)}, 2019, pp. 5565--5572.

\bibitem{hou2019learning}
S.~Hou, X.~Pan, C.~C. Loy, Z.~Wang, and D.~Lin, ``Learning a unified classifier
  incrementally via rebalancing,'' in \emph{Proceedings of the IEEE Conference
  on Computer Vision and Pattern Recognition (CVPR)}, 2019, pp. 831--839.

\bibitem{zhang2019your}
L.~Zhang, J.~Song, A.~Gao, J.~Chen, C.~Bao, and K.~Ma, ``Be your own teacher:
  Improve the performance of convolutional neural networks via self
  distillation,'' in \emph{Proceedings of the IEEE International Conference on
  Computer Vision (ICCV)}, 2019, pp. 3713--3722.

\bibitem{huang2020comprehensive}
Z.~Huang, Y.~Zou, B.~V. K.~V. Kumar, and D.~Huang, ``Comprehensive attention
  self-distillation for weakly-supervised object detection,'' in
  \emph{Proceedings of the Advances in Neural Information Processing Systems
  (NeurIPS)}, 2020, pp. 16\,797--16\,807.

\bibitem{zheng2021se}
W.~Zheng, W.~Tang, L.~Jiang, and C.-W. Fu, ``Se-ssd: Self-ensembling
  single-stage object detector from point cloud,'' in \emph{Proceedings of the
  IEEE Conference on Computer Vision and Pattern Recognition (CVPR)}, 2021, pp.
  14\,494--14\,503.

\bibitem{wang2021towards}
Y.~Wang, S.~Lin, Y.~Qu, H.~Wu, Z.~Zhang, Y.~Xie, and A.~Yao, ``Towards compact
  single image super-resolution via contrastive self-distillation,'' in
  \emph{Proceedings of the Thirtieth International Joint Conference on
  Artificial Intelligence (IJCAI)}, 2021, pp. 1122--1128.

\bibitem{shen2019amalgamating}
C.~Shen, X.~Wang, J.~Song, L.~Sun, and M.~Song, ``Amalgamating knowledge
  towards comprehensive classification,'' in \emph{Proceedings of the AAAI
  Conference on Artificial Intelligence (AAAI)}, 2019, pp. 3068--3075.

\bibitem{luo2017thinet}
J.-H. Luo, J.~Wu, and W.~Lin, ``Thinet: A filter level pruning method for deep
  neural network compression,'' in \emph{Proceedings of the IEEE International
  Conference on Computer Vision (ICCV)}, 2017, pp. 5058--5066.

\bibitem{he2017channel}
Y.~He, X.~Zhang, and J.~Sun, ``Channel pruning for accelerating very deep
  neural networks,'' in \emph{Proceedings of the IEEE International Conference
  on Computer Vision (ICCV)}, 2017, pp. 1389--1397.

\bibitem{he2016deep}
K.~He, X.~Zhang, S.~Ren, and J.~Sun, ``Deep residual learning for image
  recognition,'' in \emph{Proceedings of the IEEE Conference on Computer Vision
  and Pattern Recognition (CVPR)}, 2016, pp. 770--778.

\bibitem{krizhevsky2009learning}
A.~Krizhevsky, G.~Hinton \emph{et~al.}, ``Learning multiple layers of features
  from tiny images,'' 2009.

\bibitem{russakovsky2015imagenet}
O.~Russakovsky, J.~Deng, H.~Su, J.~Krause, S.~Satheesh, S.~Ma, Z.~Huang,
  A.~Karpathy, A.~Khosla, M.~Bernstein \emph{et~al.}, ``Imagenet large scale
  visual recognition challenge,'' \emph{International Journal of Computer
  Vision (IJCV)}, vol. 115, no.~3, pp. 211--252, 2015.

\bibitem{zagoruyko2016wide}
S.~Zagoruyko and N.~Komodakis, ``Wide residual networks,'' \emph{arXiv preprint
  arXiv:1605.07146}, 2016.

\bibitem{szegedy2015going}
C.~Szegedy, W.~Liu, Y.~Jia, P.~Sermanet, S.~Reed, D.~Anguelov, D.~Erhan,
  V.~Vanhoucke, and A.~Rabinovich, ``Going deeper with convolutions,'' in
  \emph{Proceedings of the IEEE Conference on Computer Vision and Pattern
  Recognition (CVPR)}, 2015, pp. 1--9.

\bibitem{huang2017densely}
G.~Huang, Z.~Liu, L.~Van Der~Maaten, and K.~Q. Weinberger, ``Densely connected
  convolutional networks,'' in \emph{Proceedings of the IEEE Conference on
  Computer Vision and Pattern Recognition (CVPR)}, 2017, pp. 4700--4708.

\bibitem{simonyan2014very}
K.~Simonyan and A.~Zisserman, ``Very deep convolutional networks for
  large-scale image recognition,'' \emph{arXiv preprint arXiv:1409.1556}, 2014.

\end{thebibliography}

\ifCLASSOPTIONcaptionsoff
  \newpage
\fi



%\bibliographystyle{IEEEtran}


\begin{IEEEbiography}[{\includegraphics[width=1in,height=1.25in,clip,keepaspectratio]{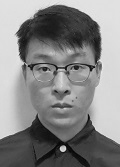}}]{Shaojie Li}
studied for his B.S. degrees in FuZhou University, China, in 2019. He is currently trying to pursue a M.S. degree in Xiamen University, China. His research interests include model compression and computer vision.
\end{IEEEbiography}

\begin{IEEEbiography}[{\includegraphics[width=1in,height=1.25in,clip,keepaspectratio]{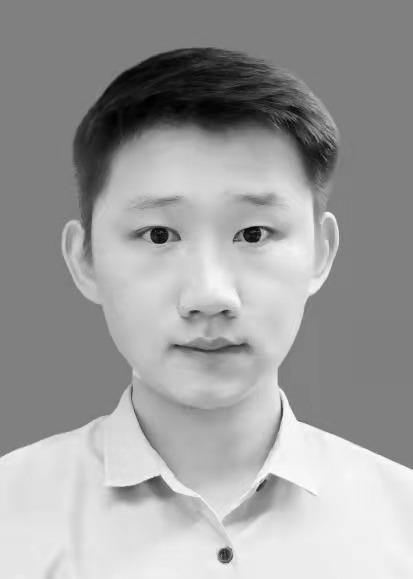}}]{Mingbao Lin} is currently pursuing the Ph.D degree with Xiamen University, China. He has published over ten papers as the first author in international journals and conferences, including IEEE TPAMI, IJCV, IEEE TIP, IEEE TNNLS, IEEE CVPR, NeurIPS, AAAI, IJCAI, ACM MM and so on. His current research interest includes network compression \& acceleration, and information retrieval.
\end{IEEEbiography}

\begin{IEEEbiography}[{\includegraphics[width=1in,height=1.25in,clip,keepaspectratio]{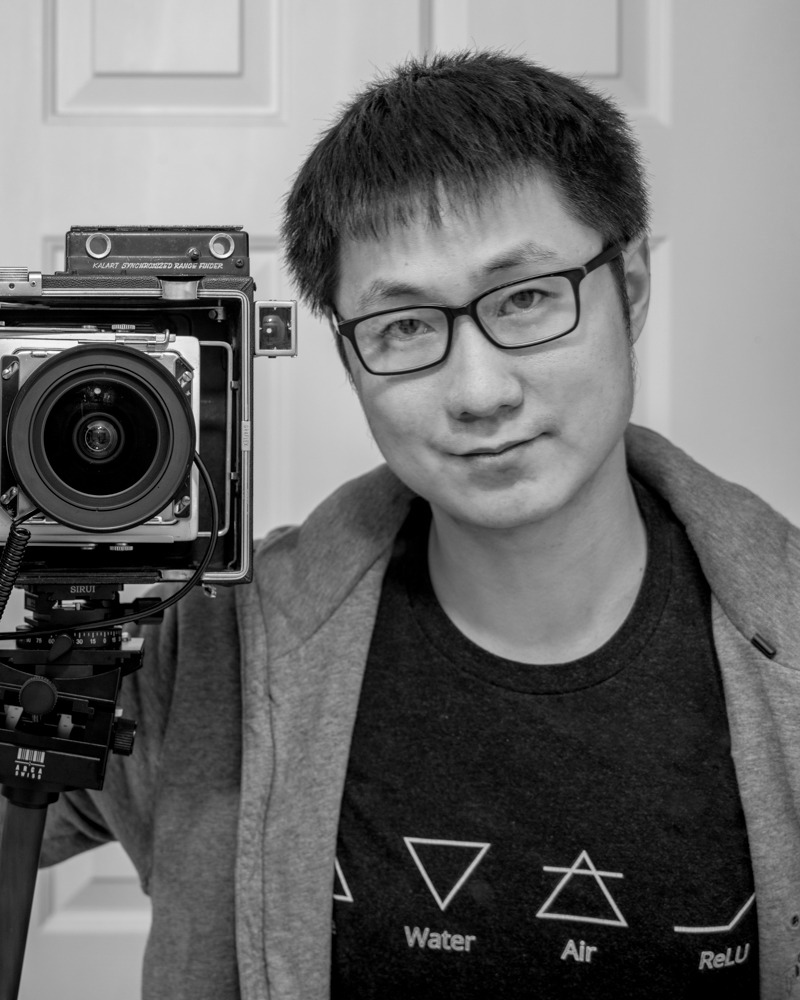}}]{Yan Wang}
works as a software engineer in Search at Pinterest. With a Ph.D degree on Electrical Engineering from Columbia University, Yan published over 20 papers on top international conferences and journals, and holds 10 US or international patents. He has broad interests on deep learning's applications on multimedia retrieval.
\end{IEEEbiography}

\begin{IEEEbiography}[{\includegraphics[width=1in,height=1.25in,clip,keepaspectratio]{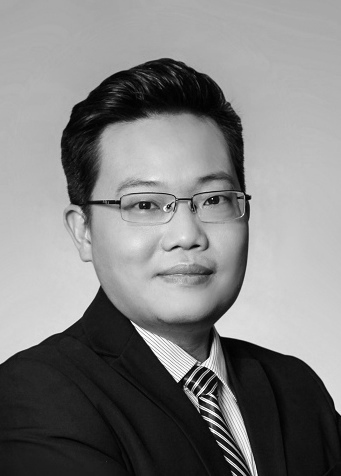}}]{Yongjian Wu} received the master degree in computer science from Wuhan University, China, in 2008. He is currently the Expert Researcher and the Deputy General Manager of Youtu Lab, Tencent Co., Ltd. His research interests include face recognition, image understanding, and large scale data processing.
\end{IEEEbiography}

\begin{IEEEbiography}[{\includegraphics[width=1in,height=1.25in,clip,keepaspectratio]{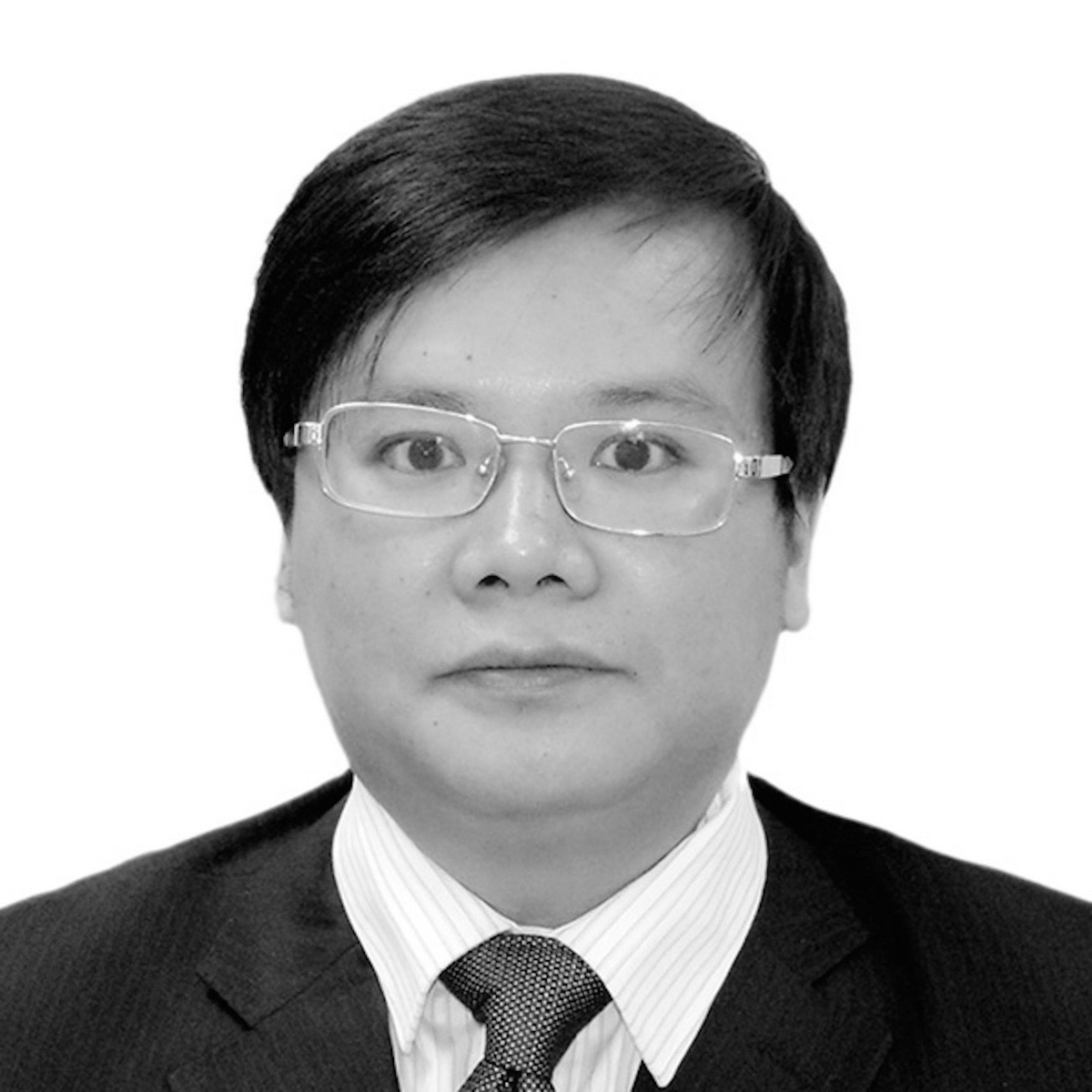}}]{Yonghong Tian} (Fellow, IEEE)
received the Ph.D. degree from the Institute of Computing Technology, Chinese Academy of Sciences, Beijing, China, in 2005. He is currently a Full Professor with the National Engineering Laboratory for Video Technology, School of Electronics Engineering and Computer Science, Peking University, Beijing, China. He has authored or coauthored more than 160 technical articles in refereed journals and conferences, and has owned more than 57 Chinese and US patents. His research interests include machine learning, computer vision, and multimedia big data.

%Prof. Tian is a Senior Member of IEEE, CIE and CCF, and a Member of ACM. He is currently an Associate Editor of the \textsc{IEEE TRANSACTIONS ON MULTIMEDIA}, \textsc{IEEE TRANSACTIONS ON CIRCUITS AND SYSTEMS FOR VIDEO TECHNOLOGY}, \textsc{IEEE MULTIMEDIA MAGAZINE}, and \textsc{IEEE ACCESS}, and a co-Editor-in-Chief of the International Journal of Multimedia Data Engineering and Management. He has served as the Technical Program Co-Chair of IEEE ICME 2015, IEEE BigMM 2015, IEEE ISM 2015 and IEEE MIPR 2018/2019, an Organizing Committee Member of more than ten conferences such as ACM Multimedia 2009, IEEE MMSP 2011, IEEE ISCAS 2013, and IEEE ISM 2016, and BigMMs 2018, and a PC Member or Area Chair of several conferences such as CVPR, ICCV, KDD, AAAI, ACM MM, ECCV, and ICME. He was the recipient of two national prizes and three ministerial prizes in China, and was the recipient of the 2015 EURASIP Best Paper Award for the EURASIP Journal on Image and Video Processing.
\end{IEEEbiography}

\begin{IEEEbiography}[{\includegraphics[width=1in,height=1.25in,clip,keepaspectratio]{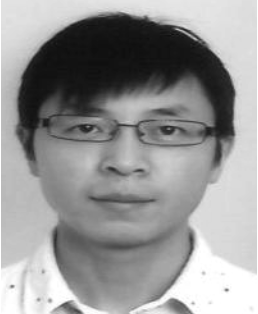}}]{Ling Shao} (Fellow, IEEE) is currently the Exec- utive Vice President and a Provost of the Mo- hamed bin Zayed University of Artificial Intelli- gence. He is also the CEO and the Chief Sci- entist of the Inception Institute of Artificial Intel- ligence (IIAI), Abu Dhabi, United Arab Emirates. His research interests include computer vision, machine learning, and medical imaging. He is a fellow of IEEE, IAPR, IET, and BCS.
\end{IEEEbiography}

\begin{IEEEbiography}[{\includegraphics[width=1in,height=1.25in,clip,keepaspectratio]{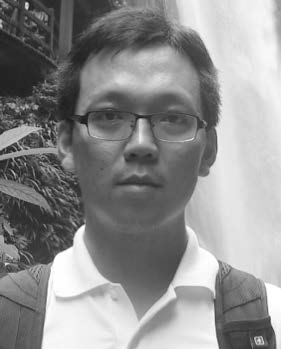}}]{Rongrong Ji}
(Senior Member, IEEE) is a Nanqiang Distinguished Professor at Xiamen University, the Deputy Director of the Office of Science and Technology at Xiamen University, and the Director of Media Analytics and Computing Lab. He was awarded as the National Science Foundation for Excellent Young Scholars (2014), the National Ten Thousand Plan for Young Top Talents (2017), and the National Science Foundation for Distinguished Young Scholars (2020). His research falls in the field of computer vision, multimedia analysis, and machine learning. He has published 50+ papers in ACM/IEEE Transactions, including TPAMI and IJCV, and 100+ full papers on top-tier conferences, such as CVPR and NeurIPS. His publications have got over 10K citations in Google Scholar. He was the recipient of the Best Paper Award of ACM Multimedia 2011. He has served as Area Chairs in top-tier conferences such as CVPR and ACM Multimedia. He is also an Advisory Member for Artificial Intelligence Construction in the Electronic Information Education Committee of the National Ministry of Education.
\end{IEEEbiography}

\end{document}